%% file: icml_paper.tex
\documentclass{article}
\pdfoutput=1
\usepackage{microtype}
\usepackage{graphicx}
\usepackage{booktabs} 

\usepackage{hyperref}

\usepackage[accepted]{icml2024}

\usepackage{amsmath}
\usepackage{amssymb}
\usepackage{bm}
\usepackage{mathtools}
\usepackage{amsthm}

\usepackage{csquotes}
\usepackage[font=small]{caption}
\usepackage[font=small]{subcaption}

\usepackage{enumitem}
\setlist{nosep}

\newcommand*{\textcite}{\citet}

\usepackage{pgfplots}    
\usepgfplotslibrary{groupplots,dateplot}
\usetikzlibrary{shapes,shapes.misc,patterns,shapes.arrows}

\usepackage[acronym,nomain]{glossaries}
\makeglossaries                   
\input{acronyms}                  

\newcommand{\refsec}[1]{Section~\ref{#1}}
\newcommand{\refssc}[1]{Section~\ref{#1}}
\newcommand{\refsssc}[1]{Section~\ref{#1}}

\theoremstyle{plain}

\theoremstyle{definition}

\theoremstyle{remark}

\usepackage[textsize=tiny]{todonotes}

\icmltitlerunning{Robustness of Explainable Artificial Intelligence}

\begin{document}

\twocolumn[
    \icmltitle{Robustness of Explainable Artificial Intelligence in Industrial Process Modelling}




    \begin{icmlauthorlist}
        \icmlauthor{Benedikt Kantz}{spsc}
        \icmlauthor{Clemens Staudinger}{voest}
        \icmlauthor{Christoph Feilmayr}{voest}
        \icmlauthor{Johannes Wachlmayr}{k1met}
        \icmlauthor{Alexander Haberl}{voest}
        \icmlauthor{Stefan Schuster}{voest}
        \icmlauthor{Franz Pernkopf}{spsc,cdlab}
        
    \end{icmlauthorlist}

    \icmlaffiliation{spsc}{Signal Processing and Speech Communication Laboratory, Technical University Graz, Graz, Austria}
    \icmlaffiliation{voest}{voestalpine Stahl GmbH, Linz, Austria}
    \icmlaffiliation{k1met}{K1-MET GmbH, Linz, Austria}
    \icmlaffiliation{cdlab}{Christian Doppler Laboratory for Dependable Intelligent Systems in Harsh Environments, Graz, Austria}


    \icmlcorrespondingauthor{Benedikt Kantz}{benedikt.kantz@tugraz.at}
    \icmlcorrespondingauthor{Franz Pernkopf}{pernkopf@tugraz.at}

    \icmlkeywords{eXplainable Artificial Intelligence, Robustness, Noise Analysis, Machine Learning, Industrial Process Modelling, Evaluation Methodology}

    \vskip 0.3in
]



\printAffiliationsAndNotice{}  

\begin{abstract}
\gls{xai} aims at providing understandable explanations of black box models. This paper evaluates current \gls{xai} methods by scoring them based on ground truth simulations and sensitivity analysis. To this end, we used an \gls{eaf} model to understand better the limits and robustness characteristics of \gls{xai} methods such as \gls{shap}, \gls{lime}, as well as \gls{ale} or \gls{sg} in a highly topical setting. These \gls{xai} methods were applied to various black-box models and then scored based on their correctness compared to the ground-truth sensitivity of the data-generating processes using a novel scoring evaluation methodology over a range of simulated additive noise.
 The resulting evaluation shows that the capability of the \gls{ml} models to capture the process accurately is, indeed, coupled with the correctness of the explainability of the underlying data-generating process. We show the differences between \gls{xai} methods in their ability to predict the true sensitivity of the modeled industrial process correctly.
\end{abstract}
\section{Introduction}
\label{intro}


\gls{ml} approaches have the power to model complex dependencies in demanding tasks such as industrial processes. However, the behavior of these industrial processes that rely on complex, non-linear interactions often needs to be fully understood. This results in the need for algorithms to understand and interpret how these \gls{ml} models arrive at specific predictions and how they might react to certain perturbances in the input.  In the last years, there has been an effort to provide explanations to the \gls{ml} model predictions using \gls{xai} \cite{Lundberg_2017, Ribeiro_2018, Alvarez_2018, Shrikumar_2017}.

Most of these works, even if they focus on the robustness and trustworthiness of the \gls{xai} method, have the shortcoming that they can only be evaluated through surrogate measures \cite{Crabbe_2023}, and the ground truth sensitivity of the evaluated datasets cannot be appropriately calculated \cite{Alvarez_2018}. Some existing approaches rather use data augmentation \cite{Sun_2020} or create measures estimating the importance of the features \cite{Yeh_2019}; further related work is provided in \refsec{background:related}. None of these systems, to the best of our knowledge, consider the ground truth sensitivity, or gradient, of the data-generating process that created the dataset. However, modeling the sensitivity to the inputs is vital to understanding the underlying process using proxy \gls{ml} models learned solely on data.

In this paper, we introduce data-driven evaluation of different \gls{xai} methods using a simulated process of an \gls{eaf} model and its ground truth sensitivity, providing insights into the actual limits and robustness properties of state-of-the-art \gls{ml} models and interpretability approaches. We propose to use a specifically generated dataset and perform perturbations to analyze this robustness empirically. Two central problems, however, arose when scoring these \gls{xai} methods and comparing them to a ground truth sensitivity:
\begin{enumerate}
    \item The feature importance scores over the feature dimensions are not within the same magnitude and range, requiring scaling \cite{Shrikumar_2017}.
    \item The relative sizes of different \gls{xai} feature importance scores are not necessarily aligned to each other \cite{Lundberg_2017, Apley_2016}, requiring normalization.
\end{enumerate}
Therefore, we introduce a novel evaluation methodology to solve these problems. We analyze how well the \gls{xai} methods explain the sensitivity of the input features, compared to a known ground truth sensitivity.

\section{Data-generating processes}
\label{methods}

This section outlines the two data generation processes: a toy dataset and the \gls{eaf} process simulation, including the process variables. These specifically generated datasets are necessary, as other datasets do not provide the ground truth effects $\mathbf{w}^*_i$ of the functions at the datapoints $\mathbf{x}^i$.

\subsection{Toy dataset}\label{methods:toy_gen}

To prove the effectiveness of our evaluation methodology, we first build a small polynomial data-generating system of the form
\begin{equation}
 f(x_1,x_2)=k_1 x_1^2+k_2 x_2^2+ k_3 x_1 x_2+ k_4 x_1+k_5 x_2+k_6
\end{equation}
with the coefficients $k_d$ being drawn from $k_d\sim\mathrm{Uniform}(0,1)$ once. The function effects, or gradients, were estimated using automatic gradient calculation using PyTorch \cite{Paszke_2019}. This data-generating process was used to generate 5000 samples $f(x_1,x_2)$, where $x_1,x_2\sim\mathrm{Uniform}(-5,5)$, which were utilized for the evaluation process.

\subsection{\gls{eaf} process}\label{methods:data_gen}

The choice of an \gls{eaf} model as the data-generating process was owed to various factors. First, the \gls{eaf} process is relevant in the steel industry as it promises, given enough clean energy, greenhouse gas emission reduction compared to traditional steel production in blast furnaces \cite{DeRas_201981}. Furthermore, \glspl{eaf} have been studied and modeled for a long time using many different approaches and modeling strategies, from their electrical characteristics \cite{Billings_1979, Boulet_2003} to their chemical processes and internal interactions \cite{Zhang1995, Basu2008}. These works helped build an accurate chemical simulation of an \gls{eaf}, representing the real-world system well while keeping it simulatable and manageable in parameter space  \cite{Aiman_2023}. The functional complexity is sophisticated, providing an interesting process to be modeled by \gls{ml} models. 

Furthermore, the data generating process is non-\gls{iid}, as the \gls{eaf} model has timesteps, and, when considering these, dependencies between each simulated furnace tapping arise, making modeling even more challenging. The model itself simulated the individual zones of the \gls{eaf} reactor as homogeneous zones, with the same temperature and uniform mixture across the whole zone. These zones are the gas zone, solid metal zone, liquid slag zone, liquid metal, and solid slag zone. Additionally, the reactor is modeled as a few discrete parts, particularly the roof and walls. The liquid metal and slag zones are the ones where measurements were simulated during tapping.

The \gls{eaf} model \cite{Aiman_2023} was converted into a Python module, where automatic differentiation tools \cite{Paszke_2019} were used to generate ground truth sensitivities of the input parameters. This simulated experiment was repeated over sampled combinations of a subset of different input parameters, which were the
oxygen lance rate,
oxygen for post-combustion,
power of the arc,
carbon injection rate,
ferromanganese injection rate, and
the mass addition rate of solids.
Furthermore, auxiliary properties within the simulated tapped material were recorded, specifically the ratios of silicon dioxide and iron oxide in the slag and the temperature of the liquid slag and metal. The observed target variable, the ratio of carbon in the tapped steel from the furnace, was recorded, too. Each tapping was considered as one data sample $\mathbf{x}_i\in \left\{ \mathbf{x}_1, \ldots, \mathbf{x}_n \right\}$, consisting of the observed variables. The gradient of the input parameters concerning the output was also calculated and accumulated at the simulation's timesteps. This simulation resulted in a dataset of about $n\approx10^4$ samples after the removal of numerical outliers due to instabilities after simulating the furnace for about a week. The process was furthermore restarted after each parameter change.

\subsection{Perturbing the Dataset}\label{methods:data_gen:perturb}

While the datasets at hand provided a perfect ground truth sensitivity for interpretations, there was still no proper way to assess the robustness of the \gls{ml} models in combination with the \gls{xai} methods. To this end, the dataset was artificially perturbed using noise in two ways. The source of the noise was Gaussian noise added to the feature $j$ of sample $i$, except the target variable, using
\begin{equation}
    \tilde{x}_{i,j}=x_{i,j}+n_j
\end{equation}
where $n_j \sim \mathcal{N}(0, l\cdot (\underset{i\in{1,\ldots,n}}{\max}~x_{i,j} - \underset{i\in{1,\ldots,n}}{~\min}~x_{i,j}))$, $x_{i,j}$ being the $j$-th feature of the observation $\mathbf{x}_i=\{x_{i,0}, x_{i,1}, \ldots, x_{i,d}\}$. The $0$-th feature is the target $y_i\in\mathcal{Y}$. $l\in [0,1)$ is the selected noise level of the experiment.

\section{Scoring Methodology for Local Explanations}\label{methods:measure}

\begin{figure*}[t]
    \centering
    \newcommand{\evalplotsize}{0.7}
    \input{figures/overview_eval.tex}
    \caption{Evaluation methodology, see text for further details and choices of measures.}
    \label{fig:graph_eval}
\end{figure*}
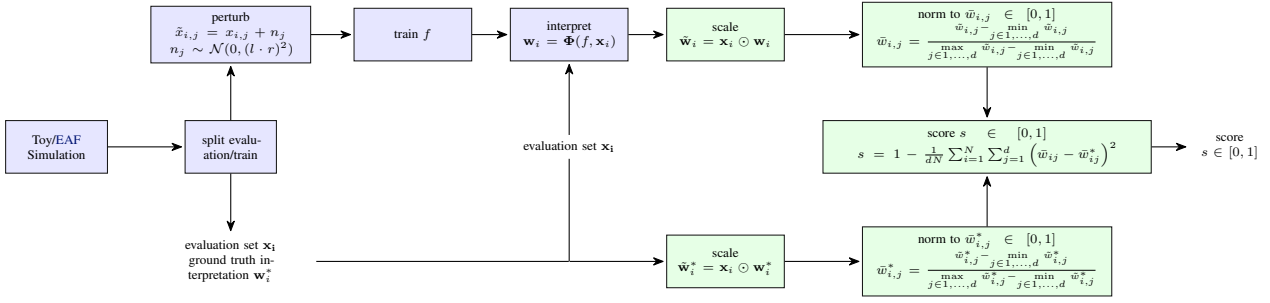

We developed an evaluation methodology (see Figure~\ref{fig:graph_eval}) to quantify the effect of the perturbances on the \gls{ml} models and the interpretation, as there is no such common measure to cater to heterogeneous types of feature importance and interpretability measures, as discussed in \refssc{background:related}. Furthermore, the problems that make comparisons difficult, as mentioned in \refsec{intro}, were addressed using this scoring methodology.

We denote the output of each \gls{xai} method as $\mathbf{w}_{i}={w_{i,1},\ldots, w_{i,d}}$, independent of the underlying explainer. All ground truth sensitivity values from the data-generating processes are denoted as $\mathbf{w}_i^*$. All input features of the data-generating functions were used in this evaluation system. To solve the first problem outlined in the introduction, the feature scores $w_{i,j}$ were scaled using the features values themselves, using
\begin{equation}
    \tilde{w}_{i,j}=w_{i,j} \cdot x_{i,j}.
\end{equation}

This scaling helps to adjust the magnitude of the \gls{xai} method results to the appropriate scale. The second problem is approached by min-max normalization on each data sample to indicate their relative strength within a sample. The scaling results in an indication of the weights' approximate strength by scaling them to a $[0,1]$ range using
\begin{equation}
    \bar{w}_{i,j}=\frac{\tilde{w}_{i,j}  - \underset{j\in{1,\ldots,d}}{~\min}~\tilde{w}_{i,j} }{\underset{j\in{1,\ldots,d}}{\max}~\tilde{w}_{i,j} - \underset{j\in{1,\ldots,d}}{~\min}~\tilde{w}_{i,j}}.
\end{equation}

Both scaling operations are performed for the ground truth derivatives $w^*_{i,j}$ of the data-generating process, too, leading to the normalized ground truth $\bar{w}^*_{i,j}$. Using both results, the score of the local interpretation can finally be computed using the Brier Score \cite{BRIER_1950}, a probabilistic scoring method initially intended for weather forecasts. It is, in essence, the \gls{mse}, and the score $s_i$ for one observation is thus calculated using
\begin{equation}
 s_i=\frac{1}{d}\sum_{j=1}^{d} \left(\bar{w}_{i,j}-\bar{w}^*_{i,j}\right)^2.
\end{equation}
The final score $s$ for one set of $n$ observations is computed by averaging each samples score $s_i$ using
\begin{equation}
 s=1-\frac{1}{n}\sum_{i=1}^{n} s_i.
\end{equation}
This averaging should return a score of $s=1$ if the interpretations completely align with the ground truth and lower if there are discrepancies. This scoring methodology, due to the comparison to the ground truth effect, focuses, therefore, not on the feature's importance but rather on the correctness of the effects that a feature has at a certain data point.

\subsection{Evaluation Process}\label{methods:eval}

The evaluation methodology is performed 50 times over a randomly sampled fold of 10 percent of the samples from the datasets. However, due to the non-iid data of the \gls{eaf}, care is taken during sampling. Therefore, the simulation runs are sampled so that no data from one run can be taken into the evaluation and training set. The sampled training set (90\% of the samples) is then perturbed using a range of different noise levels using the approaches from \refsssc{methods:data_gen:perturb}. Figure~\ref{fig:graph_eval} illustrates the whole evaluation methodology, showing how the explanations are scaled and used throughout the process. The blue part of the graph illustrates existing work and systems, while the green part shows the novel scaling and scoring scheme.
\subsubsection{Black-box Models}\label{methods:model}

The regressor \gls{ml} models $f(\mathbf{x}_i)$ used for this evaluation are a linear regression model, a neural network ensemble consisting of four networks with three layers of 32 neurons with \gls{relu} activation modeled in PyTorch \cite{Paszke_2019}, and the \gls{lgbm} regressor \cite{Ke_2017}. Explanations $\mathbf{w}_i=\bm{\Phi}(f, \mathbf{x}_i)$ are generated using the \gls{xai} methods discussed in the next section. The necessary gradients are calculated using either the parameters directly from linear regression, automatic differentiation for the neural network, or finite differences for the \gls{lgbm} tree. These local explanations are only generated on the unperturbed and complete evaluation set. This scheme allows us to test how well and accurately \gls{ml} models can learn feature importances even in noisy settings.

\subsubsection{Explanation Methods}\label{methods:xai}

The introduced score $s$ and the three \gls{ml} models $f(\mathbf{x})$ are then used in five different \gls{xai} methods. An overview of the general \gls{xai} landscape is provided in \refsec{background:iai}, showing the different types of \gls{xai} methods and why we chose local explanations. The selected local \gls{xai} methods can generally be categorized in either \textit{Effect-based Methods} (EM) and \textit{Additive Methods} (AM). The first \gls{xai} method describes the effects the input shift has on the output; the latter how much the feature contributes to the output - usually in a sparse form. The following \gls{xai} methods are used (theoretical details in \refsec{background:appr}):

\begin{itemize}
    \item The gradient baseline simply takes the gradient of the input of the \gls{ml} model - a very simple approach (EM).
    \item Improving upon that, the \gls{sg} method averages the gradient over $k=10$ neighbors sampled using \gls{knn} \cite{Yeh_2019} (EM).
    \item Next, the \gls{ale}-\gls{knn} \cite{Apley_2016} uses a simple Gaussian conditional distribution with a fixed $\sigma^2=0.2$ scaled by the feature range, a resolution of $n_{samples}=50$ bins over the feature range and $k=10$ samples for the \gls{knn} selection (EM).
    \item \gls{lime} \cite{Ribeiro_2016} uses default values, with no modifications (AM).
    \item The \gls{shap} \cite{Lundberg_2017} methods are used with default parameters. Different \gls{shap} methods, however, are used, depending on the \gls{ml} model - Tree \gls{shap} for \gls{lgbm} and Gradient \gls{shap} for the neural network ensemble and \gls{gp} (AM).
\end{itemize}

\section{Results \& Discussion}
\label{results}

This section presents the evaluation results, beginning with an overview of how noise affects the different \gls{xai} approaches in combination with different \gls{ml} models. Sanity checks with just noise are also performed, providing a lower empirical bar for $s$ of the evaluated systems.

\subsection{Robustness Results}\label{results:main}
\newcommand{\scorefigsize}{1in}
\newcommand{\scorefig}[2]{
    \begin{subfigure}[c]{\scorefigsize}
        \centering
        \vskip 0.2in
        \input{#1}
        \vskip -0.1in
        \caption{}
        \label{#2}
        \vskip -0.2in
    \end{subfigure}
}
\newcommand{\xaifigscale}{0.3}
\begin{figure*}[t]

    \begin{subfigure}[c]{\scorefigsize}
        \centering
        \vskip 0.2in
        \input{figures/noise_sweep/Polynomial_R2_Score_free.tex}
        \vskip -0.1in
        \caption{}
        \label{fig:toy_r2}
        \vskip -0.2in
    \end{subfigure}

    \hfill
    
    \begin{subfigure}[c]{\scorefigsize}
        \centering
        \vskip 0.2in
        \input{figures/noise_sweep/Polynomial_Grad_x_Input_free.tex}
        \vskip -0.1in
        \caption{}
        \label{fig:toy_grad}
        \vskip -0.2in
    \end{subfigure}

    \hfill
    
    \begin{subfigure}[c]{\scorefigsize}
        \centering
        \vskip 0.2in
        \input{figures/noise_sweep/Polynomial_SG_free.tex}
        \vskip -0.1in
        \caption{}
        \label{fig:toy_sg}
        \vskip -0.2in
    \end{subfigure}

    \hfill
    
    \begin{subfigure}[c]{\scorefigsize}
        \centering
        \vskip 0.2in
        \input{figures/noise_sweep/Polynomial_ALE_kNN_free.tex}
        \vskip -0.1in
        \caption{}
        \label{fig:toy_ale}
        \vskip -0.2in
    \end{subfigure}

    \hfill
    
    \begin{subfigure}[c]{\scorefigsize}
        \centering
        \vskip 0.2in
        \input{figures/noise_sweep/Polynomial_LIME_free.tex}
        \vskip -0.1in
        \caption{}
        \label{fig:toy_lime}
        \vskip -0.2in
    \end{subfigure}

    \hfill
    
    \begin{subfigure}[c]{\scorefigsize}
        \centering
        \vskip 0.2in
        \input{figures/noise_sweep/Polynomial_SHAP_free.tex}
        \vskip -0.1in
        \caption{}
        \label{fig:toy_shap}
        \vskip -0.2in
    \end{subfigure}

    \caption[Score for toy data with varying noise]{Score $s$ on toy data with varying levels of noise on the different combinations of explainers and \gls{ml} models. The shaded area is the 90th and 10th percentile over 50 experiments with random sampling. (\subref{fig:toy_r2}) $R^2$ score, (\subref{fig:toy_grad}) Gradient score $s$, (\subref{fig:toy_sg}) \gls{sg} score $s$, (\subref{fig:toy_ale}) \gls{ale}-\gls{knn} score $s$, (\subref{fig:toy_lime}) \gls{lime} score $s$, and (\subref{fig:toy_shap}) \gls{shap} score $s$.}
    \label{fig:noise_analy_toy}
    \vskip -0.2in
\end{figure*}
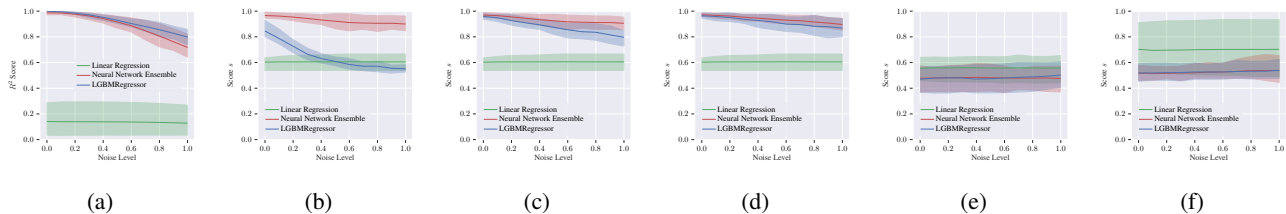
\begin{figure*}[t]
    \centering 
    
    \begin{subfigure}[c]{\scorefigsize}
        \centering
        \vskip 0.2in
        \input{figures/noise_sweep/EAF_Model_LGBMRegressor_R2_Score_free.tex}
        \vskip -0.1in
        \caption{}
        \label{fig:eaf_r2}
        \vskip -0.2in
    \end{subfigure}

    \hfill
    
    \begin{subfigure}[c]{\scorefigsize}
        \centering
        \vskip 0.2in
        \input{figures/noise_sweep/EAF_Model_LGBMRegressor_Grad_x_Input_free.tex}
        \vskip -0.1in
        \caption{}
        \label{fig:eaf_grad}
        \vskip -0.2in
    \end{subfigure}

    \hfill
    
    \begin{subfigure}[c]{\scorefigsize}
        \centering
        \vskip 0.2in
        \input{figures/noise_sweep/EAF_Model_LGBMRegressor_SG_free.tex}
        \vskip -0.1in
        \caption{}
        \label{fig:eaf_sg}
        \vskip -0.2in
    \end{subfigure}

    \hfill
    
    \begin{subfigure}[c]{\scorefigsize}
        \centering
        \vskip 0.2in
        \input{figures/noise_sweep/EAF_Model_LGBMRegressor_ALE_kNN_free.tex}
        \vskip -0.1in
        \caption{}
        \label{fig:eaf_ale}
        \vskip -0.2in
    \end{subfigure}

    \hfill
    
    \begin{subfigure}[c]{\scorefigsize}
        \centering
        \vskip 0.2in
        \input{figures/noise_sweep/EAF_Model_LGBMRegressor_SHAP_free.tex}
        \vskip -0.1in
        \caption{}
        \label{fig:eaf_shap}
        \vskip -0.2in
    \end{subfigure}

    \hfill
    
    \begin{subfigure}[c]{\scorefigsize}
        \centering
        \vskip 0.2in
        \input{figures/noise_sweep/EAF_Model_LGBMRegressor_LIME_free.tex}
        \vskip -0.1in
        \caption{}
        \label{fig:eaf_lime}
        \vskip -0.2in
    \end{subfigure}

    \caption[Score for \gls{eaf} with varying noise]{Score $s$ on \gls{eaf} data with varying levels of noise on the different combinations of explainers and the \gls{lgbm}. The shaded area is, again, the 90th and 10th percentile over 50 experiments with random sampling.  (\subref{fig:eaf_r2}) $R^2$ score, (\subref{fig:eaf_grad}) Gradient score $s$, (\subref{fig:eaf_sg}) \gls{sg} score $s$, (\subref{fig:eaf_ale}) \gls{ale}-\gls{knn} score $s$, (\subref{fig:eaf_lime}) \gls{lime} score $s$, and (\subref{fig:eaf_shap}) \gls{shap} score $s$,}
    \label{fig:noise_analy}
    \vskip -0.2in
\end{figure*}

\captionsetup{justification=centering}

The core results of this paper from both datasets are shown in Figure~\ref{fig:noise_analy_toy} and \ref{fig:noise_analy}, where we show the effect of progressively applying more noise to the training data. Various black-box models are then fitted to this perturbed training set, and \gls{xai} methods are evaluated using the scoring method proposed in \refsec{methods:measure}.

The first graphs of either dataset, Figures \ref{fig:toy_r2} and \ref{fig:eaf_r2}, show the $R^2$ score, a metric to evaluate regression problems. The score can achieve a maximum of $1$ if the predictions are perfect, $0$ if they predict the mean, and arbitrarily negative if the prediction is worse than the mean \cite{Chicco_2021}. These scores show that the increasing noise has, as expected, adverse effects on the performance of the evaluation set. The following graphs of the toy dataset, Figures \ref{fig:toy_grad} through \ref{fig:toy_ale} show that the effect-based \gls{xai} methods are strongly dependent on the performance of the trained \gls{ml} models. Linear regression fails to capture most relations while \gls{lgbm} and the neural network work quite well, especially in regimes of low noise using a robust explainer like \gls{ale}-\gls{knn}. The additive methods, shown in Figures \ref{fig:toy_lime} and \ref{fig:toy_shap}, are not able to capture the sensitivity present in the ground truth $\mathbf{w}^*$ even without noise.

Similarly, the \gls{eaf} results show that the explainer performance recorded is coupled with the \gls{ml} model performance, as the effect-based methods fall with increasing noise. This rising noise plays, again, a vital role in the lowering of the scores of the \gls{lgbm} for the more robust \gls{xai} methods, \gls{sg} and \gls{ale}-\gls{knn}, as seen in Figures \ref{fig:eaf_sg} and \ref{fig:eaf_ale}. These two and the raw gradient of Figure \ref{fig:eaf_grad} show two further interesting findings: first, when the \gls{lgbm} performance is good, the explainer score of the  \gls{lgbm} is relatively constant over the range of noise levels. Second, the \gls{lgbm} needs some initial noise on this dataset to start modeling the relations well, which can be observed for all effect-based methods, where all \gls{xai} method's scores increase when adding an initial bit of noise. This rise could be due to a regularizing effect of the noise on the gradients, effectively creating more truthful averages of gradients when adding noise. The additive methods are again not as effective in explaining the ground truth sensitivity as evident in Figures \ref{fig:eaf_lime} and \ref{fig:eaf_shap}

\subsection{Random baseline: Empirical lower bounds for Results}\label{results:sanity}

We additionally empirically calculate the random lower bounds using noisy data to ensure the evaluation measures worked. This check was performed by training the \gls{ml} model on the correct, slightly perturbed data using a noise level of $l=0.5$. The \gls{ml} model was then given random evaluation data, on which the scores for the interpreters were calculated. The evaluation dataset was set to $x_{ij}\sim\mathcal{N}(\mathbf{\mu}_j, \mathbf{\sigma}_j)$, where $\mathbf{\mu}_j$ and $\mathbf{\sigma}_j$ are the mean and standard deviation of the training dataset; the used dataset is the toy dataset. The results for this perturbation can be seen in Table~\ref{tab:results_sanity_test_noise}, where all \gls{xai} methods approached a score of about $0.5$ to $0.6$, indicating that this is indeed similar to the worst results observed above.
\begin{table*}[t]
    \caption[Results of sanity check with perturbed evaluation inputs]{Results of sanity check with Gaussian noise as evaluation inputs on the toy dataset ($\pm$ one standard deviation).}
    \label{tab:results_sanity_test_noise}
    \vskip 0.15in
    \begin{center}
        \begin{small}
            \begin{sc}
                \input{tables/sanity_xai_val.tex}
            \end{sc}
        \end{small}
    \end{center}
    \vskip -0.1in
\end{table*}
\begin{table*}[t]
    \caption[Results of sanity check with noisy training inputs]{Results of sanity check using Gaussian noise as training inputs on the toy dataset ($\pm$ one standard deviation).}
    \label{tab:results_sanity_train_noise}
    \vskip 0.15in
    \begin{center}
        \begin{small}
            \begin{sc}
                \input{tables/sanity_xai_train.tex}
            \end{sc}
        \end{small}
    \end{center}
    \vskip -0.1in
\end{table*}

In the next experiment, we trained the \gls{ml} model on completely noisy data, again with $x_{ij}\sim \mathcal{N}(\mathbf{\mu}_j, \mathbf{\sigma}_j)$, but scored on the correct evaluation set, shown in Table~\ref{tab:results_sanity_train_noise}. This produced, expectedly, even similar performance metrics. This indicates that the random baseline for this scoring method is around $s=0.55$. The neural network ensemble, however, managed to retain some performance with the effect-based explainers around $s=0.75$ with a high variability.

\subsection{Discussion}
\label{results:discussion}

This investigation of the robustness of \gls{xai} methods showed that the noise and the predictive performance influence these approaches. This is especially true for gradient-based \gls{xai} approaches. The \gls{sg} as well as the cohort-based \gls{ale}-\gls{knn} works well. The properties of the custom cohort approach foster the correctness of the interpretations, as there is an uplift in performance, especially in combination with the tree-based \gls{lgbm} system. However, both \gls{lime} and \gls{shap} are in the category of additive methods. This weakness makes them unsuitable for interpretations that call for scores reflecting the input sensitivity on the output, i.e., reflecting the feature \emph{effect} of the output.

The low performance of the additive feature importance scores can also be partially attributed to the metric preferring the effect-based scoring methods, as the related work by \textcite{Yeh_2019} notes. Notwithstanding, we showed that effect-based scoring methods are highly dependent on the performance of the \gls{ml} model to accurately reflect the ground truth importance scores $w^*_{ij}$ and that a single gradient of the \gls{ml} model is often not enough to estimate them correctly.

\section{Conclusion \& Future Work}\label{conclusion:conclusion}
We showed how different \gls{xai} methods are affected based on the predictive performance of the \gls{ml} models. This work focused on model-agnostic post-hoc explanations for local data samples, as these could be evaluated using numeric, data-driven approaches. Of these \gls{xai} methods, \gls{shap}, \gls{lime}, \gls{sg}, and a local version of \gls{ale} were chosen. These were evaluated using a novel evaluation process focused on scaling the feature importance scores to a similar magnitude within one sample, normalizing them to the same range as the ground truth effects, and calculating the distance to the ground truth effects reference. This ground truth was generated using a chemical simulation of an \gls{eaf} model, providing the necessary ground truth sensitivity for comparison and evaluation. This data-generating distribution was chosen based on the maturity of the  \gls{eaf} models for these real-world processes and current interest in the technology due to its promise of cleaner steel production. A toy example was also initially used to test the approaches on a limited and known nonlinear dataset.

Noise analysis over a range of perturbances of the initial dataset was performed using this evaluation methodology. The resulting analysis concludes that \gls{lgbm}, in combination with smooth gradient-based \gls{xai} methods, can approximate both the target values and the ground truth interpretations very well, even in noisy environments. \gls{lime} and \gls{shap} were not successfull in correctly finding the ground truth feature importance scores, probably due to their differing approaches to interpreting the feature importance scores. Furthermore, some of these \gls{xai} methods and \gls{ml} models showed a higher variance, indicating that they varied between sampling runs and were affected by the high noise.  Sanity checks on the validity of the evaluation process were carried out using noise, first, as evaluation data, and then as training data. Both tests showed that the scores are around $s=0.55$ since the \gls{ml} models cannot learn the importance at all.
\subsection{Future Work}\label{conclusion:future}

There is an apparent need to quantify the uncertainty of the \gls{xai} methods, as this high variability of the feature importance scores of  \gls{ml} models with high uncertainty in noisy environments distorted the feature interpretation significantly. There are already works investigating such uncertainties for \gls{xai} \cite{Lofstrom_2024, Zhao_2021, Slack_2021}, however, none of these address the effect-based approaches where the feature importance score reflects the change of the output with respect to the input.

The empirical evaluation of feature importance scores, especially from \gls{lime} and \gls{shap}, could also be further investigated by comparing different metrics on a ground truth dataset. Further improvements on the measures of infidelity and sensitivity \cite{Yeh_2019}, combined with the consideration of a known ground truth feature importance and deeper analysis of noise could also lead to further understanding of robustness and failure cases of \gls{xai}.

\section*{Acknowledgements}
 
The financial support by the Austrian Federal Ministry of Labour and Economy, the National Foundation for Research, Technology and Development and the Christian Doppler Research Association is gratefully acknowledged. We furthermore thankfully acknowledge the financial support of the project by voestalpine Stahl GmbH. The authors gratefully acknowledge the funding support of K1-MET GmbH, whose research program is supported by COMET (Competence Center for Excellent Technologies), the Austrian program for competence centers. COMET is funded by the Austrian ministries BMK and BMDW, the Federal States of Upper Austria, Tyrol, and Styria, and the Styrian Business Promotion Agency (SFG).

\bibliography{bibliography.bib}
\bibliographystyle{icml2024}

\newpage
\appendix
\onecolumn

\section{Background \& Related Work}
\label{background}

\subsection{Interpretable and Explainable \gls{ml}}\label{background:iai}

Before diving into the \gls{xai} methods themselves, the notion of interpretability and explainability of \gls{ml} models has to be examined first. These terms span a wide field of approaches, with the common goal of enabling the end users or creators of a system to foster their understanding on \textit{why}, and sometimes \textit{how} \gls{ml} models arrive at certain conclusions. However, the presentation of these explanations can vary widely, from visual explainers to mathematical formulas. This paper will focus on the latter, as we will take advantage of that they are easier to compare and evaluate numerically in a data-driven environment. Most methods can, despite this variety in modalities, generally be categorized into either \textit{interpretable systems} or so-called \textit{post-hoc} methods \cite{Murdoch_2019, DoshiVelez_2017}.

\subsubsection{Interpretable Systems}\label{background:iai:interpretable}

\gls{ml} models can achieve inherent interpretability through different modalities, which, however, share the constraint of understandable processes that can be comprehended by a single person through a reasonable timeframe and effort. This limitation, coined by \textcite{Murdoch_2019} as \enquote{simulatability}, greatly inhibits both the choice of system and the function space of the considered models. This leads, depending on the reduction in expressive power, to a loss of predictive performance and has to be considered before choosing such a model. The (numerical) evaluation of these systems is furthermore hindered by the wide-ranging type of approaches and often depends on human evaluation of these systems \cite{Murdoch_2019}.

One of the two common ways to increase the range of available systems again can be the introduction of sparsity, where there is an initial, more complex function and feature space that can later be reduced by fitting the model to the data and reducing the set of expressions needed for decisionmaking. Similarly, the model could be broken up into smaller modules, where each is very well understood on its own - a popular choice for these modularized, linear components would be ensemble-based approaches with only a few models \cite{Murdoch_2019}.

\subsubsection{Post-hoc Interpretability}\label{background:iai:post_hoc}

These interpretable approaches, however, do not foster the understanding of the model if there are no alternatives to the current model due to performance limitations or other constraining factors limiting choice. In these cases, post-hoc approaches can offer interpretability for black-box models that would be too complicated for humans to comprehend easily. These can be further categorized into global (dataset-level) and local (instance-level) approaches to interpretability. Global explanations are of interest for an overview of how the model considers certain features in general or for statistical analysis. If the interest lies in specific examples, or even failure cases such as adversarial examples, where a prediction can be maliciously flipped, a local explanation is often the needed approach \cite{Murdoch_2019}. Furthermore, one can also group or cluster features and analyze these cohorts with global methods, achieving a similar overview-level interpretation on a smaller neighborhood \cite{Munn_2022}. All levels utilize similar techniques to arrive at solutions explaining the decision, with the core concept being called \textit{feature scores}. These methods measure how influential a certain feature is, either on the local or global level, and assign it a score based on the perturbances it might achieve. These methods such as \textit{\glspl{pdp}} and \textit{\gls{ale}} \cite{Apley_2016} (in the 1D case), or their local counterpart, \textit{\gls{sg}}, describe how varying a single feature affects the output, with the assumption that features of interest are not correlated. We will call these methods \textit{effect-based} methods throughout this paper. \textit{Feature interactions}, on the other hand, analyze exactly these contributions: how two or more features influence the output in conjunction \cite{Naser_2021}. Other methods, such as \textit{\gls{lime}} utilize simple, local representations of models as surrogates to better explain the behavior of models in certain neighborhoods \cite{Ribeiro_2016, Munn_2022}. Both \gls{lime} and \gls{shap}, another explanation methodology, form a group of systems we will call \textit{feature importance scores}, where each feature is assigned a certain contribution to the output.

\subsection{Chosen Approaches}\label{background:appr}

Of these methods, the local and cohort-based post-hoc approaches for feature importance are of interest for the evaluation task at hand, as these allow for comparison to known feature influences from the simulated data-generating process. This is why we chose \gls{ale} on a small cohort chosen using \gls{knn} (\gls{ale}-\gls{knn}), \gls{sg}, \gls{lime} and \gls{shap}.

\subsubsection{\gls{ale}-\gls{knn} and \gls{sg}}\label{background:appr:ale_knn}

The intuition behind \gls{ale} \cite{Apley_2016} is to simply accumulate the local effects (derivatives) of a function (model) weighted by a certain prior. This prior is the main difference to \glspl{pdp}, as these do not consider the conditional distribution $p_{2|1}(x_2|x_1)$ of the sample, just the marginal over the selected variable. This results in a very intuitive and straightforward formulation of the \gls{ale} effect $f_{1,ALE}(x_1)$ on $x_1$ for a two-dimensional function ($d=2$, $f(x_1, x_2)$) as
\begin{equation}
 f_{1,ALE}(x_1)=\int_{x_{min,1}}^{x_1}\int p_{2|1}(x_2|z_1)f^1(z_1,x_2) dx_2 dz_1 - \mathrm{constant}
\end{equation}
where $f^1(x_1,x_2)$ is the first derivative with respect to the variable of interest ($\frac{\partial f(x_1,x_2)}{\partial x_1}$). $x_{min,1}$ is chosen such that it just touches the lower support of the conditional distribution, approximating the integral over the whole distribution. The first integration variable $x_2$ integrates over all samples of the second, unrelated feature (or more in real-world examples), while $z_1$ integrates the effect over the conditional distribution.  The resulting uncentered \gls{ale} is retrieved when ignoring the constant, which can be centered by subtracting the average effect. This method is usually applied on a dataset level and generates visual plots, showing the influence of variables over the whole range of the input \cite{Apley_2016}.

However, to apply this method to a local data sample, the $k$ nearest neighbors of the sample are chosen, and the resulting local cohort \cite{Munn_2022} is used as the basis for the uncentered \gls{ale} of this sample. This local cohort version of the \gls{ale} is very closely related to the \gls{sg} method, where the gradients or, more general, the explanations are averaged over a small neighborhood \cite{Yeh_2019}. The version using \glspl{ale} has, however, the advantage of the introduction of the conditional distribution, effectively weighting the effect locally.

\subsubsection{\gls{lime}}\label{background:appr:lime}

\gls{lime} \cite{Ribeiro_2016}, offers feature importance scores through local surrogate models, learning a sparse representation of the neighborhood. This approach combines almost all concepts outlined in the overview: simulatability, sparsity, local cohorts, and surrogate models. The resulting method returns feature importance scores. The algorithm itself consists of three major steps: sample selection, feature importance calculation, and a final selection process called \enquote{submodular pick}. For local explanations, only the first two steps are necessary.

Sample selection searches a training dataset for similar datapoints using a configurable distance metric, and evaluates the model for each. This small set is then used for fitting the surrogate model to the local neighborhood. A linear regressor is usually used as a local surrogate. The weights of the resulting model are then directly utilized as the feature importance scores \cite{Ribeiro_2016}.

\subsubsection{\gls{shap}}\label{background:appr:shap}

The \gls{shap} \cite{Lundberg_2017} framework unifies multiple methods for calculating \gls{shap} values, a measure of feature importance. They are based on Shapley values, the only set of values satisfying the three desirable properties for additive feature attribution. This scoring method can be described as a simple linear model of the form

\begin{equation}
 g(z') = \phi_0+\sum_{i=i}^{M} \phi_i z_i',
\end{equation}

where the model is similar to the function at $g(z') \approx f(h_x(z'))$ and $z'\in \{0,1\}^M$ being a binary representation of $x'$, a \textit{simplified input}, using the mapping function $h_x(z')=x$. The parameters $\phi_i$ describe the actual feature attribution.

While \gls{lime} follows the same structure, it does not necessarily fulfill the needed theoretical properties. \textcite{Lundberg_2017} introduce, based on this notion, three properties. These properties are
\begin{enumerate}
    \item Local accuracy: the linear model matches the complex model in the local  point;
    \item Missingness: the instances with $x_i'=0$ have no impact, so $\phi_i=0$; and
    \item Consistency: a change of the model output w.r.t. one feature resulting in an output increase should change the attribution $\phi_i$ only positively.
\end{enumerate}

The exact computation of Shapley values is quite challenging, as it is NP-hard - there are, however, heuristics and simplifications for existing models such as Kernel \gls{shap}, which is model agnostic. Other methods such as Deep \gls{shap} or Tree \gls{shap} utilize certain model-specific aspects to speed up the computation and provide more exact estimates. All of these approaches are united in the \gls{shap} package \cite{Lundberg_2017}.

\subsection{Related Work}\label{background:related}

Most of the mentioned methods prove their value either using human evaluation \cite{Ribeiro_2016, Lundberg_2017, Ribeiro_2018}, with the use of toy examples \cite{Ribeiro_2016, Apley_2016, Shrikumar_2017, Ribeiro_2018} or improvements in theoretical properties \cite{Lundberg_2017}. None of these works consider robustness to perturbances, unstable outputs, or other chaotic behavior. Some introductions of newer methods, however, are aware of these problems and decide to forego human evaluation in favor of data-driven analysis and robustness studies \cite{Sun_2020}. Further works augment this missing robustness consideration in \gls{xai}, studying these problems in more depth \cite{Alvarez_2018, Yeh_2019, Bhat_2022, Huang_2023, Crabbe_2023}. \textcite{Alvarez_2018} and \textcite{Yeh_2019} tackle these problems through the maximal local sensitivity, where they search for the maximal change of the explanation to the instance within a neighborhood. This method asses the stability of local explanations and can thus be seen as a measurement for robustness, especially on noisy data. Other methods try to reduce the sensitivity through aggregation by combining different metrics into a more resilient measure \cite{Bhat_2022}. \textcite{Yeh_2019} introduces, besides the sensitivity of the explanation, the infidelity of explainable approaches, where they measure the squared difference from a projected explanation ($\mathbf{I^\mathrm{T}}\Phi(f,\mathbf{x})$) to the difference of the output when subtracting the projection from the input ($f(\mathbf{x})-f(\mathbf{x}-\mathbf{I})$) using
\begin{equation}
    \mathrm{INFD}(\Phi, f, \mathbf{x})=
    \mathbb{E}_{\mathbf{I}\sim \mu_\mathbf{I}}\left[\left(\mathbf{I^\mathrm{T}}\Phi(f,\mathbf{x})- (f(\mathbf{x})-f(\mathbf{x}-\mathbf{I}))\right)^2\right]
\end{equation}
where $\mathbf{I} \in \mathbb{R}^d$ representing the significant perturbances around $\mathbf{x}\in \mathbb{R}^d$ as a random variable, $\mu_\mathbf{I}$ the distribution of these perturbances, $f(\mathbf{x})$ the learned model function and $\Phi(f,\mathbf{x})$ the corresponding explaination. While this measure is more data-driven and universal, it still suffers from only using an approximation of the ideal explanation. Furthermore, no studies regarding the robustness with regard to added noise or known noise levels have been performed.



\end{document}

%% file: acronyms.tex
\newacronym[shortplural=PCBs, longplural=printed circuit boards]{pcb}{PCB}{printed circuit board}
\newacronym[shortplural=IAIs, longplural=Interpretable Artifical Intelligence]{iai}{IAI}{Interpretable Artifical Intelligence}
\newacronym[shortplural=DNNs, longplural=Deep Neural Networks]{dnn}{DNN}{Deep Neural Network}
\newacronym[shortplural=NNs, longplural=Neural Networks]{nn}{NN}{Neural Network}
\newacronym[shortplural=EAFs, longplural=Electric Arc Furnaces]{eaf}{EAF}{Electric Arc Furnace}
\newacronym[shortplural=ALEs, longplural=Averaged Local Effects]{ale}{ALE}{Averaged Local Effects}
\newacronym[shortplural=PDPs, longplural=partial dependence plots]{pdp}{PDP}{Partial Dependence Plot}
\newacronym{ml}{ML}{Machine Learning}
\newacronym{ai}{AI}{Artifical Intelligence}
\newacronym{lime}{LIME}{Local Interpretable Model-agnostic Explanations}
\newacronym{knn}{kNN}{k-Nearest Neighbors}
\newacronym{shap}{SHAP}{SHapley Additive exPlanations}
\newacronym{iid}{iid.}{independant and identically distrubuted}
\newacronym{mse}{MSE}{Mean Squared Error}
\newacronym{mae}{MAE}{Mean Absolute Error}
\newacronym{relu}{ReLU}{Rectified Linear Unit}
\newacronym{xai}{XAI}{eXplainable Artificial Intelligence}
\newacronym{sg}{SG}{Smooth Gradients}
\newacronym{ig}{IG}{Integrated Gradients}
\newacronym{ofc}{OFC}{Optical Flow Constraint}
\newacronym[shortplural=GMMs, longplural=Gaussian Mixture Models]{gmm}{GMM}{Gaussian Mixture Model}
\newacronym[shortplural=GPs, longplural=Gaussian Processes]{gp}{GP}{Gaussian Process}
\newacronym{ood}{OOD}{Out-Of-Domain}
\newacronym{ucq}{UCQ}{Uncertainty Quantification}
\newacronym{lgbm}{LightGBM}{Light Gradient Boosting Machine}
\newacronym{map}{MAP}{Maximum A Posteriori}
\newacronym{kl}{KL}{Kullback-Leibler}
\newacronym{mc}{MC}{Monte-Carlo}
\newacronym{gda}{GDA}{Gaussian Discriminatory Analysis}
\newacronym{ddu}{DDU}{Deep Deterministic Uncertainty}
\newacronym{de}{DE}{Deep Ensemble}
\newacronym{clue}{CLUE}{Counterfactual Latent Uncertainty Explanations}
\newacronym{vae}{VAE}{Variational Autoencoder}
\newacronym{rbf}{RBF}{Radial Basis Function}
\newacronym[shortplural=PIs, longplural=prediction intervals]{pi}{PI}{Prediction Interval}

%% file: figures/overview_eval.tex
\usetikzlibrary{patterns,shapes.arrows}
\usetikzlibrary {arrows.meta,graphs, shapes.misc, matrix}
\begin{tikzpicture}[scale=\evalplotsize,every node/.style={scale=\evalplotsize},
    mapping/.style={rectangle, draw=black!90,  fill=blue!10, minimum size=1cm,  text width=0.9cm, font={\scriptsize}, align=center},
    mapping_nov/.style={rectangle, draw=black!90,  fill=green!10, minimum size=1cm,  text width=0.9cm, font={\scriptsize}, align=center},
    point/.style={circle,inner sep=0pt,minimum size=0pt,fill=red},
    skip loop/.style={},
    text_free/.style={text width=2cm, align=center, font={\scriptsize}},
    >={Stealth[round]},
    every new ->/.style={shorten >=1pt},]
    \matrix[row sep=0.7cm,column sep=0.5cm] {
    & \node (sample) [mapping, text width=2.8cm] {perturb \\$\tilde{x}_{i,j}=x_{i,j}+n_j$\\$ n_j\sim\mathcal{N}(0,(l\cdot r)^2)$ };
    & \node (train) [mapping, text width=2cm] {train $f$ };
    & \node (interpret) [mapping, text width=2cm] {interpret \\$\mathbf{w}_i=\bm{\Phi}(f,\mathbf{x}_i)$};
    & \node (scalet) [mapping_nov, text width=2cm] {scale \\$\tilde{\mathbf{w}}_i=\mathbf{x}_i\odot \mathbf{w}_i$};
    & \node (normt) [mapping_nov, text width=4.5cm] {norm to $\bar{w}_{i,j}\in[0,1]$ \\$\bar{w}_{i,j}=\frac{\tilde{w}_{i,j}  - \underset{j\in{1,\ldots,d}}{~\min}~\tilde{w}_{i,j} }{\underset{j\in{1,\ldots,d}}{\max}~\tilde{w}_{i,j} - \underset{j\in{1,\ldots,d}}{~\min}~\tilde{w}_{i,j}}$};\\
    \node (sim) [mapping, text width=1.7cm] {Toy/\gls{eaf} \\Simulation};
    & \node (split) [mapping, text width=1.5cm] { split evaluation/train};
    &
    & \node (eval_set) [text_free] {evaluation set $\mathbf{x_i}$};
    &
    & \node (mae_val) [mapping_nov, text width=6cm] {score $s\in[0,1]$ \\$s=1-\frac{1}{d N}\sum_{i=1}^{N}\sum_{j=1}^{d} \left( \bar{w}_{ij}-\bar{w}_{ij}^*\right)^2$};
    & \node (score) [text_free, text width=1.3cm] {score $s\in[0,1]$};\\

    & \node (p1) [text_free, text width=3cm] {evaluation set $\mathbf{x_i}$ \\ground truth interpretation $\mathbf{w}^*_i$};
    &
    & \node (p2) [point] {};
    & \node (scale_val) [mapping_nov, text width=2cm] {scale \\$\tilde{\mathbf{w}}^*_i=\mathbf{x}_i\odot \mathbf{w}^*_i$};
    & \node (norm_val) [mapping_nov, text width=4.5cm] {norm to $\bar{w}^*_{i,j}\in[0,1]$ \\$\bar{w}^*_{i,j}=\frac{\tilde{w}^*_{i,j}  - \underset{j\in{1,\ldots,d}}{~\min}~\tilde{w}^*_{i,j} }{\underset{j\in{1,\ldots,d}}{\max}~\tilde{w}^*_{i,j} - \underset{j\in{1,\ldots,d}}{~\min}~\tilde{w}^*_{i,j}}$}; \\
    };
    \graph [grow right sep, branch down=7mm, use existing nodes] {
        sim ->
        split ->
        sample ->
        train ->
        interpret ->
        scalet->
        normt->
        mae_val->
        score;
        split->
        p1--
        p2 ->
        scale_val->
        norm_val->
        mae_val;
        p2 --
        eval_set ->
        interpret;

    };
\end{tikzpicture}

%% file: figures/noise_sweep/Polynomial_R2_Score_free.tex
\begin{tikzpicture}[
scale=\xaifigscale,
every axis/.style={xtick distance=0.25, legend pos=south west, legend style={
        },
            restrict y to domain=0:1
        }
]

\definecolor{darkslategray38}{RGB}{38,38,38}
\definecolor{indianred1967882}{RGB}{196,78,82}
\definecolor{lavender234234242}{RGB}{234,234,242}
\definecolor{mediumseagreen85168104}{RGB}{85,168,104}
\definecolor{steelblue76114176}{RGB}{76,114,176}

\begin{axis}[
axis background/.style={fill=lavender234234242},
axis line style={white},
legend cell align={left},
legend style={
  fill opacity=0.8,
  draw opacity=1,
  text opacity=1,
  at={(0.91,0.5)},
  anchor=east,
  draw=none,
  fill=lavender234234242
},
tick align=outside,
tick pos=left,
x grid style={white},
xlabel=\textcolor{darkslategray38}{Noise Level},
xmajorgrids,
xmin=-0.05, xmax=1.05,
xtick style={color=darkslategray38},
xtick={-0.2,0,0.2,0.4,0.6,0.8,1,1.2},
xticklabels={
  \(\displaystyle {\ensuremath{-}0.2}\),
  \(\displaystyle {0.0}\),
  \(\displaystyle {0.2}\),
  \(\displaystyle {0.4}\),
  \(\displaystyle {0.6}\),
  \(\displaystyle {0.8}\),
  \(\displaystyle {1.0}\),
  \(\displaystyle {1.2}\)
},
y grid style={white},
ylabel=\textcolor{darkslategray38}{\(\displaystyle R^2\) Score},
ymajorgrids,
ymin=0, ymax=1,
ytick style={color=darkslategray38},
ytick={0,0.2,0.4,0.6,0.8,1},
yticklabels={
  \(\displaystyle {0.0}\),
  \(\displaystyle {0.2}\),
  \(\displaystyle {0.4}\),
  \(\displaystyle {0.6}\),
  \(\displaystyle {0.8}\),
  \(\displaystyle {1.0}\)
}
]
\path [draw=mediumseagreen85168104, fill=mediumseagreen85168104, opacity=0.3, line width=0.12pt]
(axis cs:0,0.0343022942543029)
--(axis cs:0,0.287160396575928)
--(axis cs:0.100000001490116,0.293128877878189)
--(axis cs:0.200000002980232,0.293648916482925)
--(axis cs:0.300000011920929,0.293698751926422)
--(axis cs:0.400000005960464,0.293140053749084)
--(axis cs:0.5,0.291797369718552)
--(axis cs:0.600000023841858,0.289486694335938)
--(axis cs:0.699999988079071,0.286043441295624)
--(axis cs:0.800000011920929,0.281351977586746)
--(axis cs:0.899999976158142,0.275365138053894)
--(axis cs:1,0.268114757537842)
--(axis cs:1,0.0342850685119628)
--(axis cs:1,0.0342850685119628)
--(axis cs:0.899999976158142,0.0348123073577881)
--(axis cs:0.800000011920929,0.035125720500946)
--(axis cs:0.699999988079071,0.035221767425537)
--(axis cs:0.600000023841858,0.035115498304367)
--(axis cs:0.5,0.0348433732986449)
--(axis cs:0.400000005960464,0.0344615340232849)
--(axis cs:0.300000011920929,0.0340413510799408)
--(axis cs:0.200000002980232,0.0336601138114928)
--(axis cs:0.100000001490116,0.033389139175415)
--(axis cs:0,0.0343022942543029)
--cycle;

\path [draw=indianred1967882, fill=indianred1967882, opacity=0.3, line width=0.12pt]
(axis cs:0,0.971594214439392)
--(axis cs:0,0.997371971607208)
--(axis cs:0.100000001490116,0.995920884609222)
--(axis cs:0.200000002980232,0.992138528823853)
--(axis cs:0.300000011920929,0.985654991865158)
--(axis cs:0.400000005960464,0.97497376203537)
--(axis cs:0.5,0.960276937484741)
--(axis cs:0.600000023841858,0.948186123371124)
--(axis cs:0.699999988079071,0.928131824731827)
--(axis cs:0.800000011920929,0.90278172492981)
--(axis cs:0.899999976158142,0.864215117692947)
--(axis cs:1,0.821174842119217)
--(axis cs:1,0.642639136314392)
--(axis cs:1,0.642639136314392)
--(axis cs:0.899999976158142,0.693799078464508)
--(axis cs:0.800000011920929,0.722095161676407)
--(axis cs:0.699999988079071,0.776603704690933)
--(axis cs:0.600000023841858,0.838792020082474)
--(axis cs:0.5,0.867647618055344)
--(axis cs:0.400000005960464,0.90739763379097)
--(axis cs:0.300000011920929,0.932008421421051)
--(axis cs:0.200000002980232,0.954754334688187)
--(axis cs:0.100000001490116,0.972834306955338)
--(axis cs:0,0.971594214439392)
--cycle;

\path [draw=steelblue76114176, fill=steelblue76114176, opacity=0.3, line width=0.12pt]
(axis cs:0,0.998631975653704)
--(axis cs:0,0.999452930780953)
--(axis cs:0.100000001490116,0.998107006877521)
--(axis cs:0.200000002980232,0.993474067530043)
--(axis cs:0.300000011920929,0.984662271550927)
--(axis cs:0.400000005960464,0.970110572384096)
--(axis cs:0.5,0.957831166803062)
--(axis cs:0.600000023841858,0.941234710832348)
--(axis cs:0.699999988079071,0.92963986124744)
--(axis cs:0.800000011920929,0.913160731008826)
--(axis cs:0.899999976158142,0.882452789919648)
--(axis cs:1,0.857615072111397)
--(axis cs:1,0.737451529552965)
--(axis cs:1,0.737451529552965)
--(axis cs:0.899999976158142,0.784707464369581)
--(axis cs:0.800000011920929,0.801036625579917)
--(axis cs:0.699999988079071,0.843958750118619)
--(axis cs:0.600000023841858,0.877955968572522)
--(axis cs:0.5,0.904569025152041)
--(axis cs:0.400000005960464,0.92763907004561)
--(axis cs:0.300000011920929,0.949022399188729)
--(axis cs:0.200000002980232,0.975639561241364)
--(axis cs:0.100000001490116,0.992311915715479)
--(axis cs:0,0.998631975653704)
--cycle;

\addplot [line width=0.7pt, mediumseagreen85168104]
table {%
0 0.141171185743241
0.100000001490116 0.139052578806877
0.200000002980232 0.138929679989815
0.300000011920929 0.138707941770554
0.400000005960464 0.13834193944931
0.5 0.137750658392906
0.600000023841858 0.136828017234802
0.699999988079071 0.135459470748901
0.800000011920929 0.133542221784592
0.899999976158142 0.131004306674004
1 0.127817729115486
};
\addlegendentry{Linear Regression}
\addplot [line width=0.7pt, indianred1967882]
table {%
0 0.987345028491247
0.100000001490116 0.985822531580925
0.200000002980232 0.976265770196915
0.300000011920929 0.962892439961433
0.400000005960464 0.942659172415733
0.5 0.915306240320206
0.600000023841858 0.886677038669586
0.699999988079071 0.84769335091114
0.800000011920929 0.806748789548874
0.899999976158142 0.762122577428818
1 0.718121662735939
};
\addlegendentry{Neural Network Ensemble}
\addplot [line width=0.7pt, steelblue76114176]
table {%
0 0.998926111033148
0.100000001490116 0.995432069092344
0.200000002980232 0.984856936700518
0.300000011920929 0.970990779198389
0.400000005960464 0.952252677749166
0.5 0.929001849176367
0.600000023841858 0.903766431433515
0.699999988079071 0.880459838701209
0.800000011920929 0.855727349024325
0.899999976158142 0.827861951190021
1 0.79792375022192
};
\addlegendentry{LGBMRegressor}
\end{axis}

\end{tikzpicture}

%% file: figures/noise_sweep/Polynomial_Grad_x_Input_free.tex
\begin{tikzpicture}[
scale=\xaifigscale,
every axis/.style={xtick distance=0.25, legend pos=south west, legend style={
        },
            restrict y to domain=0:1
        }
]

\definecolor{darkslategray38}{RGB}{38,38,38}
\definecolor{indianred1967882}{RGB}{196,78,82}
\definecolor{lavender234234242}{RGB}{234,234,242}
\definecolor{mediumseagreen85168104}{RGB}{85,168,104}
\definecolor{steelblue76114176}{RGB}{76,114,176}

\begin{axis}[
axis background/.style={fill=lavender234234242},
axis line style={white},
legend cell align={left},
legend style={
  fill opacity=0.8,
  draw opacity=1,
  text opacity=1,
  at={(0.03,0.03)},
  anchor=south west,
  draw=none,
  fill=lavender234234242
},
tick align=outside,
tick pos=left,
x grid style={white},
xlabel=\textcolor{darkslategray38}{Noise Level},
xmajorgrids,
xmin=-0.05, xmax=1.05,
xtick style={color=darkslategray38},
xtick={-0.2,0,0.2,0.4,0.6,0.8,1,1.2},
xticklabels={
  \(\displaystyle {\ensuremath{-}0.2}\),
  \(\displaystyle {0.0}\),
  \(\displaystyle {0.2}\),
  \(\displaystyle {0.4}\),
  \(\displaystyle {0.6}\),
  \(\displaystyle {0.8}\),
  \(\displaystyle {1.0}\),
  \(\displaystyle {1.2}\)
},
y grid style={white},
ylabel=\textcolor{darkslategray38}{Score \(\displaystyle s\)},
ymajorgrids,
ymin=0, ymax=1,
ytick style={color=darkslategray38},
ytick={0,0.2,0.4,0.6,0.8,1},
yticklabels={
  \(\displaystyle {0.0}\),
  \(\displaystyle {0.2}\),
  \(\displaystyle {0.4}\),
  \(\displaystyle {0.6}\),
  \(\displaystyle {0.8}\),
  \(\displaystyle {1.0}\)
}
]
\path [draw=mediumseagreen85168104, fill=mediumseagreen85168104, opacity=0.3, line width=0.12pt]
(axis cs:0,0.53600001335144)
--(axis cs:0,0.636000037193298)
--(axis cs:0.100000001490116,0.647800004482269)
--(axis cs:0.200000002980232,0.654999965429306)
--(axis cs:0.300000011920929,0.656799995899201)
--(axis cs:0.400000005960464,0.658800023794174)
--(axis cs:0.5,0.664400011301041)
--(axis cs:0.600000023841858,0.666200041770935)
--(axis cs:0.699999988079071,0.666200041770935)
--(axis cs:0.800000011920929,0.666000038385391)
--(axis cs:0.899999976158142,0.666000038385391)
--(axis cs:1,0.666200041770935)
--(axis cs:1,0.535600012540817)
--(axis cs:1,0.535600012540817)
--(axis cs:0.899999976158142,0.535600012540817)
--(axis cs:0.800000011920929,0.535600012540817)
--(axis cs:0.699999988079071,0.539199966192246)
--(axis cs:0.600000023841858,0.539199966192246)
--(axis cs:0.5,0.539399969577789)
--(axis cs:0.400000005960464,0.53600001335144)
--(axis cs:0.300000011920929,0.537799990177155)
--(axis cs:0.200000002980232,0.539399969577789)
--(axis cs:0.100000001490116,0.539399969577789)
--(axis cs:0,0.53600001335144)
--cycle;

\path [draw=indianred1967882, fill=indianred1967882, opacity=0.3, line width=0.12pt]
(axis cs:0,0.945999979972839)
--(axis cs:0,0.994000017642975)
--(axis cs:0.100000001490116,0.986599999666214)
--(axis cs:0.200000002980232,0.983199995756149)
--(axis cs:0.300000011920929,0.978399986028671)
--(axis cs:0.400000005960464,0.973200005292893)
--(axis cs:0.5,0.980000019073486)
--(axis cs:0.600000023841858,0.980200016498566)
--(axis cs:0.699999988079071,0.967600017786026)
--(axis cs:0.800000011920929,0.964599990844727)
--(axis cs:0.899999976158142,0.964599990844727)
--(axis cs:1,0.962000012397766)
--(axis cs:1,0.847399991750717)
--(axis cs:1,0.847399991750717)
--(axis cs:0.899999976158142,0.859400010108948)
--(axis cs:0.800000011920929,0.842000007629395)
--(axis cs:0.699999988079071,0.858800011873245)
--(axis cs:0.600000023841858,0.846999990940094)
--(axis cs:0.5,0.863399964570999)
--(axis cs:0.400000005960464,0.8966000020504)
--(axis cs:0.300000011920929,0.909799969196319)
--(axis cs:0.200000002980232,0.925800001621246)
--(axis cs:0.100000001490116,0.940600025653839)
--(axis cs:0,0.945999979972839)
--cycle;

\path [draw=steelblue76114176, fill=steelblue76114176, opacity=0.3, line width=0.12pt]
(axis cs:0,0.801999986171722)
--(axis cs:0,0.927999973297119)
--(axis cs:0.100000001490116,0.860499972105026)
--(axis cs:0.200000002980232,0.778400009870529)
--(axis cs:0.300000011920929,0.713099998235702)
--(axis cs:0.400000005960464,0.67009996175766)
--(axis cs:0.5,0.643400007486343)
--(axis cs:0.600000023841858,0.631100034713745)
--(axis cs:0.699999988079071,0.602600002288818)
--(axis cs:0.800000011920929,0.611400002241135)
--(axis cs:0.899999976158142,0.57960000038147)
--(axis cs:1,0.581899976730347)
--(axis cs:1,0.526300007104874)
--(axis cs:1,0.526300007104874)
--(axis cs:0.899999976158142,0.517700004577637)
--(axis cs:0.800000011920929,0.527900040149689)
--(axis cs:0.699999988079071,0.53600001335144)
--(axis cs:0.600000023841858,0.547699975967407)
--(axis cs:0.5,0.574499988555908)
--(axis cs:0.400000005960464,0.597299987077713)
--(axis cs:0.300000011920929,0.623500007390976)
--(axis cs:0.200000002980232,0.677103328704834)
--(axis cs:0.100000001490116,0.745600014925003)
--(axis cs:0,0.801999986171722)
--cycle;

\addplot [line width=0.7pt, mediumseagreen85168104]
table {%
0 0.60257143066043
0.100000001490116 0.604400005936623
0.200000002980232 0.604299998283386
0.300000011920929 0.605099999904633
0.400000005960464 0.605000001192093
0.5 0.605100002884865
0.600000023841858 0.605500003695488
0.699999988079071 0.605100005865097
0.800000011920929 0.604900002479553
0.899999976158142 0.604600003361702
1 0.605000004172325
};
\addlegendentry{Linear Regression}
\addplot [line width=0.7pt, indianred1967882]
table {%
0 0.966000003474099
0.100000001490116 0.962100005149841
0.200000002980232 0.954100006818771
0.300000011920929 0.943299993872643
0.400000005960464 0.931599998474121
0.5 0.921700000762939
0.600000023841858 0.911700001358986
0.699999988079071 0.907600000500679
0.800000011920929 0.906000006198883
0.899999976158142 0.905899995565414
1 0.90009999871254
};
\addlegendentry{Neural Network Ensemble}
\addplot [line width=0.7pt, steelblue76114176]
table {%
0 0.845523805845351
0.100000001490116 0.787449997663498
0.200000002980232 0.72680167555809
0.300000011920929 0.665950003266335
0.400000005960464 0.631199991703033
0.5 0.609150004386902
0.600000023841858 0.584549999237061
0.699999988079071 0.571299996972084
0.800000011920929 0.571700006723404
0.899999976158142 0.555000001192093
1 0.551350000500679
};
\addlegendentry{LGBMRegressor}
\end{axis}

\end{tikzpicture}

%% file: figures/noise_sweep/Polynomial_SG_free.tex
\begin{tikzpicture}[
scale=\xaifigscale,
every axis/.style={xtick distance=0.25, legend pos=south west, legend style={
        },
            restrict y to domain=0:1
        }
]

\definecolor{darkslategray38}{RGB}{38,38,38}
\definecolor{indianred1967882}{RGB}{196,78,82}
\definecolor{lavender234234242}{RGB}{234,234,242}
\definecolor{mediumseagreen85168104}{RGB}{85,168,104}
\definecolor{steelblue76114176}{RGB}{76,114,176}

\begin{axis}[
axis background/.style={fill=lavender234234242},
axis line style={white},
legend cell align={left},
legend style={
  fill opacity=0.8,
  draw opacity=1,
  text opacity=1,
  at={(0.03,0.03)},
  anchor=south west,
  draw=none,
  fill=lavender234234242
},
tick align=outside,
tick pos=left,
x grid style={white},
xlabel=\textcolor{darkslategray38}{Noise Level},
xmajorgrids,
xmin=-0.05, xmax=1.05,
xtick style={color=darkslategray38},
xtick={-0.2,0,0.2,0.4,0.6,0.8,1,1.2},
xticklabels={
  \(\displaystyle {\ensuremath{-}0.2}\),
  \(\displaystyle {0.0}\),
  \(\displaystyle {0.2}\),
  \(\displaystyle {0.4}\),
  \(\displaystyle {0.6}\),
  \(\displaystyle {0.8}\),
  \(\displaystyle {1.0}\),
  \(\displaystyle {1.2}\)
},
y grid style={white},
ylabel=\textcolor{darkslategray38}{Score \(\displaystyle s\)},
ymajorgrids,
ymin=0, ymax=1,
ytick style={color=darkslategray38},
ytick={0,0.2,0.4,0.6,0.8,1},
yticklabels={
  \(\displaystyle {0.0}\),
  \(\displaystyle {0.2}\),
  \(\displaystyle {0.4}\),
  \(\displaystyle {0.6}\),
  \(\displaystyle {0.8}\),
  \(\displaystyle {1.0}\)
}
]
\path [draw=mediumseagreen85168104, fill=mediumseagreen85168104, opacity=0.3, line width=0.12pt]
(axis cs:0,0.53600001335144)
--(axis cs:0,0.636000037193298)
--(axis cs:0.100000001490116,0.647800004482269)
--(axis cs:0.200000002980232,0.654999965429306)
--(axis cs:0.300000011920929,0.656799995899201)
--(axis cs:0.400000005960464,0.658800023794174)
--(axis cs:0.5,0.664400011301041)
--(axis cs:0.600000023841858,0.666200041770935)
--(axis cs:0.699999988079071,0.666200041770935)
--(axis cs:0.800000011920929,0.666000038385391)
--(axis cs:0.899999976158142,0.666000038385391)
--(axis cs:1,0.666200041770935)
--(axis cs:1,0.535600012540817)
--(axis cs:1,0.535600012540817)
--(axis cs:0.899999976158142,0.535600012540817)
--(axis cs:0.800000011920929,0.535600012540817)
--(axis cs:0.699999988079071,0.539199966192246)
--(axis cs:0.600000023841858,0.539199966192246)
--(axis cs:0.5,0.539399969577789)
--(axis cs:0.400000005960464,0.53600001335144)
--(axis cs:0.300000011920929,0.537799990177155)
--(axis cs:0.200000002980232,0.539399969577789)
--(axis cs:0.100000001490116,0.539399969577789)
--(axis cs:0,0.53600001335144)
--cycle;

\path [draw=indianred1967882, fill=indianred1967882, opacity=0.3, line width=0.12pt]
(axis cs:0,0.952000021934509)
--(axis cs:0,0.990000009536743)
--(axis cs:0.100000001490116,0.984400022029877)
--(axis cs:0.200000002980232,0.97720000743866)
--(axis cs:0.300000011920929,0.978399986028671)
--(axis cs:0.400000005960464,0.97419998049736)
--(axis cs:0.5,0.974999982118607)
--(axis cs:0.600000023841858,0.969599997997284)
--(axis cs:0.699999988079071,0.966199988126755)
--(axis cs:0.800000011920929,0.962800014019012)
--(axis cs:0.899999976158142,0.962800014019012)
--(axis cs:1,0.953000020980835)
--(axis cs:1,0.857799983024597)
--(axis cs:1,0.857799983024597)
--(axis cs:0.899999976158142,0.859600013494492)
--(axis cs:0.800000011920929,0.859200012683868)
--(axis cs:0.699999988079071,0.867000025510788)
--(axis cs:0.600000023841858,0.855800002813339)
--(axis cs:0.5,0.871799981594086)
--(axis cs:0.400000005960464,0.889799988269806)
--(axis cs:0.300000011920929,0.911799997091293)
--(axis cs:0.200000002980232,0.929600006341934)
--(axis cs:0.100000001490116,0.945999979972839)
--(axis cs:0,0.952000021934509)
--cycle;

\path [draw=steelblue76114176, fill=steelblue76114176, opacity=0.3, line width=0.12pt]
(axis cs:0,0.938000023365021)
--(axis cs:0,0.980000019073486)
--(axis cs:0.100000001490116,0.972600001096726)
--(axis cs:0.200000002980232,0.972400003671646)
--(axis cs:0.300000011920929,0.953200018405914)
--(axis cs:0.400000005960464,0.93940002322197)
--(axis cs:0.5,0.924199992418289)
--(axis cs:0.600000023841858,0.905200010538101)
--(axis cs:0.699999988079071,0.905399984121323)
--(axis cs:0.800000011920929,0.900999975204468)
--(axis cs:0.899999976158142,0.886400002241135)
--(axis cs:1,0.849400013685226)
--(axis cs:1,0.727799987792969)
--(axis cs:1,0.727799987792969)
--(axis cs:0.899999976158142,0.748399972915649)
--(axis cs:0.800000011920929,0.773599988222122)
--(axis cs:0.699999988079071,0.785000014305115)
--(axis cs:0.600000023841858,0.791599994897842)
--(axis cs:0.5,0.81939999461174)
--(axis cs:0.400000005960464,0.859000015258789)
--(axis cs:0.300000011920929,0.871799981594086)
--(axis cs:0.200000002980232,0.891400015354156)
--(axis cs:0.100000001490116,0.919600015878677)
--(axis cs:0,0.938000023365021)
--cycle;

\addplot [line width=0.7pt, mediumseagreen85168104]
table {%
0 0.60257143066043
0.100000001490116 0.604400005936623
0.200000002980232 0.604299998283386
0.300000011920929 0.605099999904633
0.400000005960464 0.605000001192093
0.5 0.605100002884865
0.600000023841858 0.605500003695488
0.699999988079071 0.605100005865097
0.800000011920929 0.604900002479553
0.899999976158142 0.604600003361702
1 0.605000004172325
};
\addlegendentry{Linear Regression}
\addplot [line width=0.7pt, indianred1967882]
table {%
0 0.969714298134758
0.100000001490116 0.965499997138977
0.200000002980232 0.955300003290176
0.300000011920929 0.944499996304512
0.400000005960464 0.933699998259544
0.5 0.924999997019768
0.600000023841858 0.916899999976158
0.699999988079071 0.914400002360344
0.800000011920929 0.911100006103516
0.899999976158142 0.911300006508827
1 0.905799999833107
};
\addlegendentry{Neural Network Ensemble}
\addplot [line width=0.7pt, steelblue76114176]
table {%
0 0.959047626881372
0.100000001490116 0.948600000143051
0.200000002980232 0.926500001549721
0.300000011920929 0.909400001168251
0.400000005960464 0.893900007009506
0.5 0.874199995398521
0.600000023841858 0.855200001597405
0.699999988079071 0.840199998021126
0.800000011920929 0.835999995470047
0.899999976158142 0.815599992871284
1 0.796499997377396
};
\addlegendentry{LGBMRegressor}
\end{axis}

\end{tikzpicture}

%% file: figures/noise_sweep/Polynomial_ALE_kNN_free.tex
\begin{tikzpicture}[
scale=\xaifigscale,
every axis/.style={xtick distance=0.25, legend pos=south west, legend style={
        },
            restrict y to domain=0:1
        }
]

\definecolor{darkslategray38}{RGB}{38,38,38}
\definecolor{indianred1967882}{RGB}{196,78,82}
\definecolor{lavender234234242}{RGB}{234,234,242}
\definecolor{mediumseagreen85168104}{RGB}{85,168,104}
\definecolor{steelblue76114176}{RGB}{76,114,176}

\begin{axis}[
axis background/.style={fill=lavender234234242},
axis line style={white},
legend cell align={left},
legend style={
  fill opacity=0.8,
  draw opacity=1,
  text opacity=1,
  at={(0.03,0.03)},
  anchor=south west,
  draw=none,
  fill=lavender234234242
},
tick align=outside,
tick pos=left,
x grid style={white},
xlabel=\textcolor{darkslategray38}{Noise Level},
xmajorgrids,
xmin=-0.05, xmax=1.05,
xtick style={color=darkslategray38},
xtick={-0.2,0,0.2,0.4,0.6,0.8,1,1.2},
xticklabels={
  \(\displaystyle {\ensuremath{-}0.2}\),
  \(\displaystyle {0.0}\),
  \(\displaystyle {0.2}\),
  \(\displaystyle {0.4}\),
  \(\displaystyle {0.6}\),
  \(\displaystyle {0.8}\),
  \(\displaystyle {1.0}\),
  \(\displaystyle {1.2}\)
},
y grid style={white},
ylabel=\textcolor{darkslategray38}{Score \(\displaystyle s\)},
ymajorgrids,
ymin=0, ymax=1,
ytick style={color=darkslategray38},
ytick={0,0.2,0.4,0.6,0.8,1},
yticklabels={
  \(\displaystyle {0.0}\),
  \(\displaystyle {0.2}\),
  \(\displaystyle {0.4}\),
  \(\displaystyle {0.6}\),
  \(\displaystyle {0.8}\),
  \(\displaystyle {1.0}\)
}
]
\path [draw=mediumseagreen85168104, fill=mediumseagreen85168104, opacity=0.3, line width=0.12pt]
(axis cs:0,0.53600001335144)
--(axis cs:0,0.636000037193298)
--(axis cs:0.100000001490116,0.647800004482269)
--(axis cs:0.200000002980232,0.654999965429306)
--(axis cs:0.300000011920929,0.656799995899201)
--(axis cs:0.400000005960464,0.658800023794174)
--(axis cs:0.5,0.664400011301041)
--(axis cs:0.600000023841858,0.666200041770935)
--(axis cs:0.699999988079071,0.666200041770935)
--(axis cs:0.800000011920929,0.666000038385391)
--(axis cs:0.899999976158142,0.666000038385391)
--(axis cs:1,0.666200041770935)
--(axis cs:1,0.535600012540817)
--(axis cs:1,0.535600012540817)
--(axis cs:0.899999976158142,0.535600012540817)
--(axis cs:0.800000011920929,0.535600012540817)
--(axis cs:0.699999988079071,0.539199966192246)
--(axis cs:0.600000023841858,0.539199966192246)
--(axis cs:0.5,0.539399969577789)
--(axis cs:0.400000005960464,0.53600001335144)
--(axis cs:0.300000011920929,0.537799990177155)
--(axis cs:0.200000002980232,0.539399969577789)
--(axis cs:0.100000001490116,0.539399969577789)
--(axis cs:0,0.53600001335144)
--cycle;

\path [draw=indianred1967882, fill=indianred1967882, opacity=0.3, line width=0.12pt]
(axis cs:0,0.959999978542328)
--(axis cs:0,0.986000001430512)
--(axis cs:0.100000001490116,0.984400022029877)
--(axis cs:0.200000002980232,0.982399994134903)
--(axis cs:0.300000011920929,0.976800012588501)
--(axis cs:0.400000005960464,0.976800012588501)
--(axis cs:0.5,0.971200025081635)
--(axis cs:0.600000023841858,0.970600026845932)
--(axis cs:0.699999988079071,0.966200017929077)
--(axis cs:0.800000011920929,0.966200017929077)
--(axis cs:0.899999976158142,0.950599986314774)
--(axis cs:1,0.944800007343292)
--(axis cs:1,0.852399975061417)
--(axis cs:1,0.852399975061417)
--(axis cs:0.899999976158142,0.866800028085709)
--(axis cs:0.800000011920929,0.880799973011017)
--(axis cs:0.699999988079071,0.895800024271011)
--(axis cs:0.600000023841858,0.901000010967255)
--(axis cs:0.5,0.908399969339371)
--(axis cs:0.400000005960464,0.922800022363663)
--(axis cs:0.300000011920929,0.923600023984909)
--(axis cs:0.200000002980232,0.939999997615814)
--(axis cs:0.100000001490116,0.945599985122681)
--(axis cs:0,0.959999978542328)
--cycle;

\path [draw=steelblue76114176, fill=steelblue76114176, opacity=0.3, line width=0.12pt]
(axis cs:0,0.945999979972839)
--(axis cs:0,0.986000001430512)
--(axis cs:0.100000001490116,0.982199996709824)
--(axis cs:0.200000002980232,0.988199979066849)
--(axis cs:0.300000011920929,0.971400028467178)
--(axis cs:0.400000005960464,0.972600001096726)
--(axis cs:0.5,0.976400011777878)
--(axis cs:0.600000023841858,0.956599974632263)
--(axis cs:0.699999988079071,0.954399996995926)
--(axis cs:0.800000011920929,0.966400021314621)
--(axis cs:0.899999976158142,0.950199991464615)
--(axis cs:1,0.942200028896332)
--(axis cs:1,0.801799988746643)
--(axis cs:1,0.801799988746643)
--(axis cs:0.899999976158142,0.791599994897842)
--(axis cs:0.800000011920929,0.812000036239624)
--(axis cs:0.699999988079071,0.835599970817566)
--(axis cs:0.600000023841858,0.84239998459816)
--(axis cs:0.5,0.857399982213974)
--(axis cs:0.400000005960464,0.873200011253357)
--(axis cs:0.300000011920929,0.905200016498566)
--(axis cs:0.200000002980232,0.919400012493133)
--(axis cs:0.100000001490116,0.925400000810623)
--(axis cs:0,0.945999979972839)
--cycle;

\addplot [line width=0.7pt, mediumseagreen85168104]
table {%
0 0.60257143066043
0.100000001490116 0.604400005936623
0.200000002980232 0.604299998283386
0.300000011920929 0.605099999904633
0.400000005960464 0.605000001192093
0.5 0.605100002884865
0.600000023841858 0.605500003695488
0.699999988079071 0.605100005865097
0.800000011920929 0.604900002479553
0.899999976158142 0.604600003361702
1 0.605000004172325
};
\addlegendentry{Linear Regression}
\addplot [line width=0.7pt, indianred1967882]
table {%
0 0.972380953175681
0.100000001490116 0.967200005054474
0.200000002980232 0.962199997901916
0.300000011920929 0.95280000269413
0.400000005960464 0.945800006389618
0.5 0.937199994921684
0.600000023841858 0.930100002884865
0.699999988079071 0.925700002908707
0.800000011920929 0.915900000929832
0.899999976158142 0.907299995422363
1 0.89626966714859
};
\addlegendentry{Neural Network Ensemble}
\addplot [line width=0.7pt, steelblue76114176]
table {%
0 0.963904758294423
0.100000001490116 0.958900007605553
0.200000002980232 0.951200005412102
0.300000011920929 0.93690000474453
0.400000005960464 0.927499997615814
0.5 0.917200002074242
0.600000023841858 0.899500000476837
0.699999988079071 0.895699998736382
0.800000011920929 0.882700002193451
0.899999976158142 0.878799992799759
1 0.868500000238419
};
\addlegendentry{LGBMRegressor}
\end{axis}

\end{tikzpicture}

%% file: figures/noise_sweep/Polynomial_LIME_free.tex
\begin{tikzpicture}[
scale=\xaifigscale,
every axis/.style={xtick distance=0.25, legend pos=south west, legend style={
        },
            restrict y to domain=0:1
        }
]

\definecolor{darkslategray38}{RGB}{38,38,38}
\definecolor{indianred1967882}{RGB}{196,78,82}
\definecolor{lavender234234242}{RGB}{234,234,242}
\definecolor{mediumseagreen85168104}{RGB}{85,168,104}
\definecolor{steelblue76114176}{RGB}{76,114,176}

\begin{axis}[
axis background/.style={fill=lavender234234242},
axis line style={white},
legend cell align={left},
legend style={fill opacity=0.8, draw opacity=1, text opacity=1, draw=none, fill=lavender234234242},
tick align=outside,
tick pos=left,
x grid style={white},
xlabel=\textcolor{darkslategray38}{Noise Level},
xmajorgrids,
xmin=-0.05, xmax=1.05,
xtick style={color=darkslategray38},
xtick={-0.2,0,0.2,0.4,0.6,0.8,1,1.2},
xticklabels={
  \(\displaystyle {\ensuremath{-}0.2}\),
  \(\displaystyle {0.0}\),
  \(\displaystyle {0.2}\),
  \(\displaystyle {0.4}\),
  \(\displaystyle {0.6}\),
  \(\displaystyle {0.8}\),
  \(\displaystyle {1.0}\),
  \(\displaystyle {1.2}\)
},
y grid style={white},
ylabel=\textcolor{darkslategray38}{Score \(\displaystyle s\)},
ymajorgrids,
ymin=0, ymax=1,
ytick style={color=darkslategray38},
ytick={0,0.2,0.4,0.6,0.8,1},
yticklabels={
  \(\displaystyle {0.0}\),
  \(\displaystyle {0.2}\),
  \(\displaystyle {0.4}\),
  \(\displaystyle {0.6}\),
  \(\displaystyle {0.8}\),
  \(\displaystyle {1.0}\)
}
]
\path [draw=mediumseagreen85168104, fill=mediumseagreen85168104, opacity=0.3, line width=0.12pt]
(axis cs:0,0.459999978542328)
--(axis cs:0,0.639999985694885)
--(axis cs:0.100000001490116,0.642999964952469)
--(axis cs:0.200000002980232,0.640399986505508)
--(axis cs:0.300000011920929,0.64439999461174)
--(axis cs:0.400000005960464,0.646600025892258)
--(axis cs:0.5,0.64459999203682)
--(axis cs:0.600000023841858,0.64439999461174)
--(axis cs:0.699999988079071,0.658199989795685)
--(axis cs:0.800000011920929,0.648000001907349)
--(axis cs:0.899999976158142,0.647000026702881)
--(axis cs:1,0.654000043869019)
--(axis cs:1,0.4432000041008)
--(axis cs:1,0.4432000041008)
--(axis cs:0.899999976158142,0.441399973630905)
--(axis cs:0.800000011920929,0.439599996805191)
--(axis cs:0.699999988079071,0.447400009632111)
--(axis cs:0.600000023841858,0.437600022554398)
--(axis cs:0.5,0.447600013017654)
--(axis cs:0.400000005960464,0.447600013017654)
--(axis cs:0.300000011920929,0.451800018548965)
--(axis cs:0.200000002980232,0.448000013828278)
--(axis cs:0.100000001490116,0.454199975728989)
--(axis cs:0,0.459999978542328)
--cycle;

\path [draw=indianred1967882, fill=indianred1967882, opacity=0.3, line width=0.12pt]
(axis cs:0,0.371999979019165)
--(axis cs:0,0.574000000953674)
--(axis cs:0.100000001490116,0.562000006437302)
--(axis cs:0.200000002980232,0.576199978590012)
--(axis cs:0.300000011920929,0.579800003767014)
--(axis cs:0.400000005960464,0.592000007629395)
--(axis cs:0.5,0.588400000333786)
--(axis cs:0.600000023841858,0.573399990797043)
--(axis cs:0.699999988079071,0.546000027656555)
--(axis cs:0.800000011920929,0.570799958705902)
--(axis cs:0.899999976158142,0.570399963855743)
--(axis cs:1,0.57119996547699)
--(axis cs:1,0.371199983358383)
--(axis cs:1,0.371199983358383)
--(axis cs:0.899999976158142,0.374000012874603)
--(axis cs:0.800000011920929,0.384999978542328)
--(axis cs:0.699999988079071,0.390600001811981)
--(axis cs:0.600000023841858,0.366999989748001)
--(axis cs:0.5,0.378800016641617)
--(axis cs:0.400000005960464,0.375400012731552)
--(axis cs:0.300000011920929,0.374200004339218)
--(axis cs:0.200000002980232,0.372799998521805)
--(axis cs:0.100000001490116,0.372599995136261)
--(axis cs:0,0.371999979019165)
--cycle;

\path [draw=steelblue76114176, fill=steelblue76114176, opacity=0.3, line width=0.12pt]
(axis cs:0,0.367999970912933)
--(axis cs:0,0.564000010490417)
--(axis cs:0.100000001490116,0.568599963188171)
--(axis cs:0.200000002980232,0.572800028324127)
--(axis cs:0.300000011920929,0.575199997425079)
--(axis cs:0.400000005960464,0.584999990463257)
--(axis cs:0.5,0.584799993038178)
--(axis cs:0.600000023841858,0.589399999380112)
--(axis cs:0.699999988079071,0.600399988889694)
--(axis cs:0.800000011920929,0.598799991607666)
--(axis cs:0.899999976158142,0.59799998998642)
--(axis cs:1,0.604999971389771)
--(axis cs:1,0.408399999141693)
--(axis cs:1,0.408399999141693)
--(axis cs:0.899999976158142,0.39180001616478)
--(axis cs:0.800000011920929,0.374800020456314)
--(axis cs:0.699999988079071,0.369999986886978)
--(axis cs:0.600000023841858,0.365400010347366)
--(axis cs:0.5,0.364000010490417)
--(axis cs:0.400000005960464,0.363399988412857)
--(axis cs:0.300000011920929,0.372200000286102)
--(axis cs:0.200000002980232,0.362599980831146)
--(axis cs:0.100000001490116,0.362200003862381)
--(axis cs:0,0.367999970912933)
--cycle;

\addplot [line width=0.7pt, mediumseagreen85168104]
table {%
0 0.555100631146204
0.100000001490116 0.558700001239777
0.200000002980232 0.558399999141693
0.300000011920929 0.559100005030632
0.400000005960464 0.558700010180473
0.5 0.556799998879433
0.600000023841858 0.556800004839897
0.699999988079071 0.558399999141693
0.800000011920929 0.557200008630753
0.899999976158142 0.556599995493889
1 0.557400009036064
};
\addlegendentry{Linear Regression}
\addplot [line width=0.7pt, indianred1967882]
table {%
0 0.477809519994827
0.100000001490116 0.47940000295639
0.200000002980232 0.48170000910759
0.300000011920929 0.482800003886223
0.400000005960464 0.484700000286102
0.5 0.482000002264977
0.600000023841858 0.480100002884865
0.699999988079071 0.478300005197525
0.800000011920929 0.478699994087219
0.899999976158142 0.479600003361702
1 0.475999996066093
};
\addlegendentry{Neural Network Ensemble}
\addplot [line width=0.7pt, steelblue76114176]
table {%
0 0.4708571434021
0.100000001490116 0.479100009799004
0.200000002980232 0.48120000064373
0.300000011920929 0.479599997401237
0.400000005960464 0.471400001645088
0.5 0.474299994111061
0.600000023841858 0.479100000858307
0.699999988079071 0.485099995136261
0.800000011920929 0.488199999928474
0.899999976158142 0.493299996852875
1 0.502599993348122
};
\addlegendentry{LGBMRegressor}
\end{axis}

\end{tikzpicture}

%% file: figures/noise_sweep/Polynomial_SHAP_free.tex
\begin{tikzpicture}[
scale=\xaifigscale,
every axis/.style={xtick distance=0.25, legend pos=south west, legend style={
        },
            restrict y to domain=0:1
        }
]

\definecolor{darkslategray38}{RGB}{38,38,38}
\definecolor{indianred1967882}{RGB}{196,78,82}
\definecolor{lavender234234242}{RGB}{234,234,242}
\definecolor{mediumseagreen85168104}{RGB}{85,168,104}
\definecolor{steelblue76114176}{RGB}{76,114,176}

\begin{axis}[
axis background/.style={fill=lavender234234242},
axis line style={white},
legend cell align={left},
legend style={fill opacity=0.8, draw opacity=1, text opacity=1, draw=none, fill=lavender234234242},
tick align=outside,
tick pos=left,
x grid style={white},
xlabel=\textcolor{darkslategray38}{Noise Level},
xmajorgrids,
xmin=-0.05, xmax=1.05,
xtick style={color=darkslategray38},
xtick={-0.2,0,0.2,0.4,0.6,0.8,1,1.2},
xticklabels={
  \(\displaystyle {\ensuremath{-}0.2}\),
  \(\displaystyle {0.0}\),
  \(\displaystyle {0.2}\),
  \(\displaystyle {0.4}\),
  \(\displaystyle {0.6}\),
  \(\displaystyle {0.8}\),
  \(\displaystyle {1.0}\),
  \(\displaystyle {1.2}\)
},
y grid style={white},
ylabel=\textcolor{darkslategray38}{Score \(\displaystyle s\)},
ymajorgrids,
ymin=0, ymax=1,
ytick style={color=darkslategray38},
ytick={0,0.2,0.4,0.6,0.8,1},
yticklabels={
  \(\displaystyle {0.0}\),
  \(\displaystyle {0.2}\),
  \(\displaystyle {0.4}\),
  \(\displaystyle {0.6}\),
  \(\displaystyle {0.8}\),
  \(\displaystyle {1.0}\)
}
]
\path [draw=mediumseagreen85168104, fill=mediumseagreen85168104, opacity=0.3, line width=0.12pt]
(axis cs:0,0.51800000667572)
--(axis cs:0,0.909999966621399)
--(axis cs:0.100000001490116,0.917400008440018)
--(axis cs:0.200000002980232,0.924400019645691)
--(axis cs:0.300000011920929,0.924800020456314)
--(axis cs:0.400000005960464,0.927600002288818)
--(axis cs:0.5,0.931200009584427)
--(axis cs:0.600000023841858,0.935200011730194)
--(axis cs:0.699999988079071,0.935400015115738)
--(axis cs:0.800000011920929,0.933599984645843)
--(axis cs:0.899999976158142,0.933599984645843)
--(axis cs:1,0.9333999812603)
--(axis cs:1,0.488200002908707)
--(axis cs:1,0.488200002908707)
--(axis cs:0.899999976158142,0.490399974584579)
--(axis cs:0.800000011920929,0.492800009250641)
--(axis cs:0.699999988079071,0.500199967622757)
--(axis cs:0.600000023841858,0.50059996843338)
--(axis cs:0.5,0.502799999713898)
--(axis cs:0.400000005960464,0.504600030183792)
--(axis cs:0.300000011920929,0.5083999812603)
--(axis cs:0.200000002980232,0.514000022411346)
--(axis cs:0.100000001490116,0.512199991941452)
--(axis cs:0,0.51800000667572)
--cycle;

\path [draw=indianred1967882, fill=indianred1967882, opacity=0.3, line width=0.12pt]
(axis cs:0,0.462)
--(axis cs:0,0.578)
--(axis cs:0.100000001490116,0.5694)
--(axis cs:0.200000002980232,0.566)
--(axis cs:0.300000011920929,0.5726)
--(axis cs:0.400000005960464,0.5904)
--(axis cs:0.5,0.6)
--(axis cs:0.600000023841858,0.5956)
--(axis cs:0.699999988079071,0.626)
--(axis cs:0.800000011920929,0.6474)
--(axis cs:0.899999976158142,0.662)
--(axis cs:1,0.6554)
--(axis cs:1,0.4454)
--(axis cs:1,0.4454)
--(axis cs:0.899999976158142,0.4606)
--(axis cs:0.800000011920929,0.477)
--(axis cs:0.699999988079071,0.4782)
--(axis cs:0.600000023841858,0.4774)
--(axis cs:0.5,0.478)
--(axis cs:0.400000005960464,0.4776)
--(axis cs:0.300000011920929,0.4604)
--(axis cs:0.200000002980232,0.4732)
--(axis cs:0.100000001490116,0.4624)
--(axis cs:0,0.462)
--cycle;

\path [draw=steelblue76114176, fill=steelblue76114176, opacity=0.3, line width=0.12pt]
(axis cs:0,0.454)
--(axis cs:0,0.59)
--(axis cs:0.100000001490116,0.5952)
--(axis cs:0.200000002980232,0.5952)
--(axis cs:0.300000011920929,0.5892)
--(axis cs:0.400000005960464,0.5962)
--(axis cs:0.5,0.587)
--(axis cs:0.600000023841858,0.5964)
--(axis cs:0.699999988079071,0.616)
--(axis cs:0.800000011920929,0.6092)
--(axis cs:0.899999976158142,0.6106)
--(axis cs:1,0.6254)
--(axis cs:1,0.4752)
--(axis cs:1,0.4752)
--(axis cs:0.899999976158142,0.4782)
--(axis cs:0.800000011920929,0.4718)
--(axis cs:0.699999988079071,0.4778)
--(axis cs:0.600000023841858,0.47)
--(axis cs:0.5,0.4696)
--(axis cs:0.400000005960464,0.47)
--(axis cs:0.300000011920929,0.4572)
--(axis cs:0.200000002980232,0.462)
--(axis cs:0.100000001490116,0.4626)
--(axis cs:0,0.454)
--cycle;

\addplot [line width=0.7pt, mediumseagreen85168104]
table {%
0 0.703238089879354
0.100000001490116 0.694500002264977
0.200000002980232 0.696400001645088
0.300000011920929 0.697399997711182
0.400000005960464 0.699800002574921
0.5 0.700900000333786
0.600000023841858 0.702100005745888
0.699999988079071 0.702099993824959
0.800000011920929 0.702299997210503
0.899999976158142 0.702299991250038
1 0.702699995040894
};
\addlegendentry{Linear Regression}
\addplot [line width=0.7pt, indianred1967882]
table {%
0 0.517047619047619
0.100000001490116 0.5157
0.200000002980232 0.5163
0.300000011920929 0.5189
0.400000005960464 0.5203
0.5 0.5256
0.600000023841858 0.5264
0.699999988079071 0.5308
0.800000011920929 0.5366
0.899999976158142 0.5376
1 0.5388
};
\addlegendentry{Neural Network Ensemble}
\addplot [line width=0.7pt, steelblue76114176]
table {%
0 0.519047619047619
0.100000001490116 0.518
0.200000002980232 0.5216
0.300000011920929 0.5221
0.400000005960464 0.5266
0.5 0.5263
0.600000023841858 0.5268
0.699999988079071 0.5304
0.800000011920929 0.5329
0.899999976158142 0.5329
1 0.539
};
\addlegendentry{LGBMRegressor}
\end{axis}

\end{tikzpicture}

%% file: figures/noise_sweep/EAF_Model_LGBMRegressor_R2_Score_free.tex
\begin{tikzpicture}[
scale=\xaifigscale,
every axis/.style={xtick distance=0.25, legend pos=south west, legend style={
                           font=\Large
                },
            restrict y to domain=0:1
                }
]

\definecolor{darkslategray38}{RGB}{38,38,38}
\definecolor{lavender234234242}{RGB}{234,234,242}
\definecolor{steelblue76114176}{RGB}{76,114,176}

\begin{axis}[
axis background/.style={fill=lavender234234242},
axis line style={white},
legend cell align={left},
legend style={fill opacity=0.8, draw opacity=1, text opacity=1, draw=none, fill=lavender234234242},
tick align=outside,
tick pos=left,
x grid style={white},
xlabel=\textcolor{darkslategray38}{Noise Level},
xmajorgrids,
xmin=-0.05, xmax=1.05,
xtick style={color=darkslategray38},
xtick={-0.2,0,0.2,0.4,0.6,0.8,1,1.2},
xticklabels={
  \(\displaystyle {\ensuremath{-}0.2}\),
  \(\displaystyle {0.0}\),
  \(\displaystyle {0.2}\),
  \(\displaystyle {0.4}\),
  \(\displaystyle {0.6}\),
  \(\displaystyle {0.8}\),
  \(\displaystyle {1.0}\),
  \(\displaystyle {1.2}\)
},
y grid style={white},
ylabel=\textcolor{darkslategray38}{\(\displaystyle R^2\) Score},
ymajorgrids,
ymin=0, ymax=1,
ytick style={color=darkslategray38},
ytick={0,0.2,0.4,0.6,0.8,1},
yticklabels={
  \(\displaystyle {0.0}\),
  \(\displaystyle {0.2}\),
  \(\displaystyle {0.4}\),
  \(\displaystyle {0.6}\),
  \(\displaystyle {0.8}\),
  \(\displaystyle {1.0}\)
}
]
\path [draw=steelblue76114176, fill=steelblue76114176, opacity=0.3, line width=0.12pt]
(axis cs:0,0.982353709663415)
--(axis cs:0,0.99287964467612)
--(axis cs:0.100000001490116,0.949884737247027)
--(axis cs:0.200000002980232,0.925879775177112)
--(axis cs:0.300000011920929,0.913128493967034)
--(axis cs:0.400000005960464,0.901924129476566)
--(axis cs:0.5,0.892302877550615)
--(axis cs:0.600000023841858,0.874828004003024)
--(axis cs:0.699999988079071,0.866040224129006)
--(axis cs:0.800000011920929,0.860010922670991)
--(axis cs:0.899999976158142,0.838880049924724)
--(axis cs:1,0.827907269607829)
--(axis cs:1,0.743832044246574)
--(axis cs:1,0.743832044246574)
--(axis cs:0.899999976158142,0.757410478386167)
--(axis cs:0.800000011920929,0.789965808805956)
--(axis cs:0.699999988079071,0.805456544078473)
--(axis cs:0.600000023841858,0.826908840748981)
--(axis cs:0.5,0.84783507194207)
--(axis cs:0.400000005960464,0.865234252138726)
--(axis cs:0.300000011920929,0.876294924991059)
--(axis cs:0.200000002980232,0.901277228307702)
--(axis cs:0.100000001490116,0.921135498726572)
--(axis cs:0,0.982353709663415)
--cycle;

\addplot [line width=0.7pt, steelblue76114176]
table {%
0 0.988212181150949
0.100000001490116 0.934650789046943
0.200000002980232 0.912342541540381
0.300000011920929 0.89655563649807
0.400000005960464 0.884854771872033
0.5 0.871283135837385
0.600000023841858 0.855050222606043
0.699999988079071 0.84004438299984
0.800000011920929 0.826331307604034
0.899999976158142 0.804665367905422
1 0.786381464310116
};
\addlegendentry{LGBMRegressor}
\end{axis}

\end{tikzpicture}

%% file: figures/noise_sweep/EAF_Model_LGBMRegressor_Grad_x_Input_free.tex
\begin{tikzpicture}[
scale=\xaifigscale,
every axis/.style={xtick distance=0.25, legend pos=south west, legend style={
                           font=\Large
                },
            restrict y to domain=0:1
                }
]

\definecolor{darkslategray38}{RGB}{38,38,38}
\definecolor{lavender234234242}{RGB}{234,234,242}
\definecolor{steelblue76114176}{RGB}{76,114,176}

\begin{axis}[
axis background/.style={fill=lavender234234242},
axis line style={white},
legend cell align={left},
legend style={fill opacity=0.8, draw opacity=1, text opacity=1, draw=none, fill=lavender234234242},
tick align=outside,
tick pos=left,
x grid style={white},
xlabel=\textcolor{darkslategray38}{Noise Level},
xmajorgrids,
xmin=-0.05, xmax=1.05,
xtick style={color=darkslategray38},
xtick={-0.2,0,0.2,0.4,0.6,0.8,1,1.2},
xticklabels={
  \(\displaystyle {\ensuremath{-}0.2}\),
  \(\displaystyle {0.0}\),
  \(\displaystyle {0.2}\),
  \(\displaystyle {0.4}\),
  \(\displaystyle {0.6}\),
  \(\displaystyle {0.8}\),
  \(\displaystyle {1.0}\),
  \(\displaystyle {1.2}\)
},
y grid style={white},
ylabel=\textcolor{darkslategray38}{Score \(\displaystyle s\)},
ymajorgrids,
ymin=0, ymax=1,
ytick style={color=darkslategray38},
ytick={0,0.2,0.4,0.6,0.8,1},
yticklabels={
  \(\displaystyle {0.0}\),
  \(\displaystyle {0.2}\),
  \(\displaystyle {0.4}\),
  \(\displaystyle {0.6}\),
  \(\displaystyle {0.8}\),
  \(\displaystyle {1.0}\)
}
]
\path [draw=steelblue76114176, fill=steelblue76114176, opacity=0.3, line width=0.12pt]
(axis cs:0,0.421188235282898)
--(axis cs:0,0.436879515647888)
--(axis cs:0.100000001490116,0.58798593878746)
--(axis cs:0.200000002980232,0.566371071338654)
--(axis cs:0.300000011920929,0.595366895198822)
--(axis cs:0.400000005960464,0.568851917982101)
--(axis cs:0.5,0.584116488695145)
--(axis cs:0.600000023841858,0.570607018470764)
--(axis cs:0.699999988079071,0.572884601354599)
--(axis cs:0.800000011920929,0.551469027996063)
--(axis cs:0.899999976158142,0.581786471605301)
--(axis cs:1,0.538640850782394)
--(axis cs:1,0.45793269276619)
--(axis cs:1,0.45793269276619)
--(axis cs:0.899999976158142,0.462335193157196)
--(axis cs:0.800000011920929,0.493233966827393)
--(axis cs:0.699999988079071,0.483644288778305)
--(axis cs:0.600000023841858,0.512214106321335)
--(axis cs:0.5,0.498818224668503)
--(axis cs:0.400000005960464,0.501006233692169)
--(axis cs:0.300000011920929,0.521349412202835)
--(axis cs:0.200000002980232,0.516813051700592)
--(axis cs:0.100000001490116,0.530113953351975)
--(axis cs:0,0.421188235282898)
--cycle;

\addplot [line width=0.7pt, steelblue76114176]
table {%
0 0.430020278408414
0.100000001490116 0.55721874833107
0.200000002980232 0.541991871595383
0.300000011920929 0.55483755171299
0.400000005960464 0.539840030670166
0.5 0.536786898970604
0.600000023841858 0.539785236120224
0.699999988079071 0.519249898195267
0.800000011920929 0.518911722302437
0.899999976158142 0.516365823149681
1 0.491881608963013
};
\addlegendentry{LGBMRegressor}
\end{axis}

\end{tikzpicture}

%% file: figures/noise_sweep/EAF_Model_LGBMRegressor_SG_free.tex
\begin{tikzpicture}[
scale=\xaifigscale,
every axis/.style={xtick distance=0.25, legend pos=south west, legend style={
                           font=\Large
                },
            restrict y to domain=0:1
                }
]

\definecolor{darkslategray38}{RGB}{38,38,38}
\definecolor{lavender234234242}{RGB}{234,234,242}
\definecolor{steelblue76114176}{RGB}{76,114,176}

\begin{axis}[
axis background/.style={fill=lavender234234242},
axis line style={white},
legend cell align={left},
legend style={fill opacity=0.8, draw opacity=1, text opacity=1, draw=none, fill=lavender234234242},
tick align=outside,
tick pos=left,
x grid style={white},
xlabel=\textcolor{darkslategray38}{Noise Level},
xmajorgrids,
xmin=-0.05, xmax=1.05,
xtick style={color=darkslategray38},
xtick={-0.2,0,0.2,0.4,0.6,0.8,1,1.2},
xticklabels={
  \(\displaystyle {\ensuremath{-}0.2}\),
  \(\displaystyle {0.0}\),
  \(\displaystyle {0.2}\),
  \(\displaystyle {0.4}\),
  \(\displaystyle {0.6}\),
  \(\displaystyle {0.8}\),
  \(\displaystyle {1.0}\),
  \(\displaystyle {1.2}\)
},
y grid style={white},
ylabel=\textcolor{darkslategray38}{Score \(\displaystyle s\)},
ymajorgrids,
ymin=0, ymax=1,
ytick style={color=darkslategray38},
ytick={0,0.2,0.4,0.6,0.8,1},
yticklabels={
  \(\displaystyle {0.0}\),
  \(\displaystyle {0.2}\),
  \(\displaystyle {0.4}\),
  \(\displaystyle {0.6}\),
  \(\displaystyle {0.8}\),
  \(\displaystyle {1.0}\)
}
]
\path [draw=steelblue76114176, fill=steelblue76114176, opacity=0.3, line width=0.12pt]
(axis cs:0,0.592792749404907)
--(axis cs:0,0.634063124656677)
--(axis cs:0.100000001490116,0.698861336708069)
--(axis cs:0.200000002980232,0.704571974277496)
--(axis cs:0.300000011920929,0.717128294706345)
--(axis cs:0.400000005960464,0.737299978733063)
--(axis cs:0.5,0.723737037181854)
--(axis cs:0.600000023841858,0.716207766532898)
--(axis cs:0.699999988079071,0.697126221656799)
--(axis cs:0.800000011920929,0.685302513837814)
--(axis cs:0.899999976158142,0.69283527135849)
--(axis cs:1,0.71389764547348)
--(axis cs:1,0.581481868028641)
--(axis cs:1,0.581481868028641)
--(axis cs:0.899999976158142,0.544680190086365)
--(axis cs:0.800000011920929,0.590451866388321)
--(axis cs:0.699999988079071,0.638342094421387)
--(axis cs:0.600000023841858,0.615831363201141)
--(axis cs:0.5,0.652176457643509)
--(axis cs:0.400000005960464,0.602513897418976)
--(axis cs:0.300000011920929,0.623966068029404)
--(axis cs:0.200000002980232,0.63699791431427)
--(axis cs:0.100000001490116,0.653407949209213)
--(axis cs:0,0.592792749404907)
--cycle;

\addplot [line width=0.7pt, steelblue76114176]
table {%
0 0.612920006116231
0.100000001490116 0.673023876547813
0.200000002980232 0.667305415868759
0.300000011920929 0.67571684718132
0.400000005960464 0.664838796854019
0.5 0.684980058670044
0.600000023841858 0.674263891577721
0.699999988079071 0.660671648383141
0.800000011920929 0.639866000413895
0.899999976158142 0.637987053394318
1 0.645809653401375
};
\addlegendentry{LGBMRegressor}
\end{axis}

\end{tikzpicture}

%% file: figures/noise_sweep/EAF_Model_LGBMRegressor_ALE_kNN_free.tex
\begin{tikzpicture}[
scale=\xaifigscale,
every axis/.style={xtick distance=0.25, legend pos=south west, legend style={
                           font=\Large
                },
            restrict y to domain=0:1
                }
]

\definecolor{darkslategray38}{RGB}{38,38,38}
\definecolor{lavender234234242}{RGB}{234,234,242}
\definecolor{steelblue76114176}{RGB}{76,114,176}

\begin{axis}[
axis background/.style={fill=lavender234234242},
axis line style={white},
legend cell align={left},
legend style={fill opacity=0.8, draw opacity=1, text opacity=1, draw=none, fill=lavender234234242},
tick align=outside,
tick pos=left,
x grid style={white},
xlabel=\textcolor{darkslategray38}{Noise Level},
xmajorgrids,
xmin=-0.05, xmax=1.05,
xtick style={color=darkslategray38},
xtick={-0.2,0,0.2,0.4,0.6,0.8,1,1.2},
xticklabels={
  \(\displaystyle {\ensuremath{-}0.2}\),
  \(\displaystyle {0.0}\),
  \(\displaystyle {0.2}\),
  \(\displaystyle {0.4}\),
  \(\displaystyle {0.6}\),
  \(\displaystyle {0.8}\),
  \(\displaystyle {1.0}\),
  \(\displaystyle {1.2}\)
},
y grid style={white},
ylabel=\textcolor{darkslategray38}{Score \(\displaystyle s\)},
ymajorgrids,
ymin=0, ymax=1,
ytick style={color=darkslategray38},
ytick={0,0.2,0.4,0.6,0.8,1},
yticklabels={
  \(\displaystyle {0.0}\),
  \(\displaystyle {0.2}\),
  \(\displaystyle {0.4}\),
  \(\displaystyle {0.6}\),
  \(\displaystyle {0.8}\),
  \(\displaystyle {1.0}\)
}
]
\path [draw=steelblue76114176, fill=steelblue76114176, opacity=0.3, line width=0.12pt]
(axis cs:0,0.74256831407547)
--(axis cs:0,0.795812487602234)
--(axis cs:0.100000001490116,0.839844781160355)
--(axis cs:0.200000002980232,0.795474171638489)
--(axis cs:0.300000011920929,0.78288808465004)
--(axis cs:0.400000005960464,0.788921189308166)
--(axis cs:0.5,0.793966472148895)
--(axis cs:0.600000023841858,0.802653431892395)
--(axis cs:0.699999988079071,0.779791754484177)
--(axis cs:0.800000011920929,0.786971962451935)
--(axis cs:0.899999976158142,0.76838681101799)
--(axis cs:1,0.766661989688873)
--(axis cs:1,0.637274670600891)
--(axis cs:1,0.637274670600891)
--(axis cs:0.899999976158142,0.700944834947586)
--(axis cs:0.800000011920929,0.671944129467011)
--(axis cs:0.699999988079071,0.711624866724014)
--(axis cs:0.600000023841858,0.724114519357681)
--(axis cs:0.5,0.737791991233826)
--(axis cs:0.400000005960464,0.720665496587753)
--(axis cs:0.300000011920929,0.743389630317688)
--(axis cs:0.200000002980232,0.751722568273544)
--(axis cs:0.100000001490116,0.780108445882797)
--(axis cs:0,0.74256831407547)
--cycle;

\addplot [line width=0.7pt, steelblue76114176]
table {%
0 0.773451450325194
0.100000001490116 0.815126088261604
0.200000002980232 0.774826148152351
0.300000011920929 0.76661396920681
0.400000005960464 0.757180669903755
0.5 0.765663829445839
0.600000023841858 0.762670058012009
0.699999988079071 0.741316276788712
0.800000011920929 0.732429134845734
0.899999976158142 0.732886773347855
1 0.708457323908806
};
\addlegendentry{LGBMRegressor}
\end{axis}

\end{tikzpicture}

%% file: figures/noise_sweep/EAF_Model_LGBMRegressor_SHAP_free.tex
\begin{tikzpicture}[
scale=\xaifigscale,
every axis/.style={xtick distance=0.25, legend pos=south west, legend style={
                           font=\Large
                },
            restrict y to domain=0:1
                }
]

\definecolor{darkslategray38}{RGB}{38,38,38}
\definecolor{lavender234234242}{RGB}{234,234,242}
\definecolor{steelblue76114176}{RGB}{76,114,176}

\begin{axis}[
axis background/.style={fill=lavender234234242},
axis line style={white},
legend cell align={left},
legend style={fill opacity=0.8, draw opacity=1, text opacity=1, draw=none, fill=lavender234234242},
tick align=outside,
tick pos=left,
x grid style={white},
xlabel=\textcolor{darkslategray38}{Noise Level},
xmajorgrids,
xmin=-0.05, xmax=1.05,
xtick style={color=darkslategray38},
xtick={-0.2,0,0.2,0.4,0.6,0.8,1,1.2},
xticklabels={
  \(\displaystyle {\ensuremath{-}0.2}\),
  \(\displaystyle {0.0}\),
  \(\displaystyle {0.2}\),
  \(\displaystyle {0.4}\),
  \(\displaystyle {0.6}\),
  \(\displaystyle {0.8}\),
  \(\displaystyle {1.0}\),
  \(\displaystyle {1.2}\)
},
y grid style={white},
ylabel=\textcolor{darkslategray38}{Score \(\displaystyle s\)},
ymajorgrids,
ymin=0, ymax=1,
ytick style={color=darkslategray38},
ytick={0,0.2,0.4,0.6,0.8,1},
yticklabels={
  \(\displaystyle {0.0}\),
  \(\displaystyle {0.2}\),
  \(\displaystyle {0.4}\),
  \(\displaystyle {0.6}\),
  \(\displaystyle {0.8}\),
  \(\displaystyle {1.0}\)
}
]
\path [draw=steelblue76114176, fill=steelblue76114176, opacity=0.3, line width=0.12pt]
(axis cs:0,0.493694824786895)
--(axis cs:0,0.513380601770639)
--(axis cs:0.100000001490116,0.515526886477607)
--(axis cs:0.200000002980232,0.513383082607468)
--(axis cs:0.300000011920929,0.516103410655737)
--(axis cs:0.400000005960464,0.519708675679575)
--(axis cs:0.5,0.524252393137844)
--(axis cs:0.600000023841858,0.524537721323782)
--(axis cs:0.699999988079071,0.52821401707796)
--(axis cs:0.800000011920929,0.527441518414116)
--(axis cs:0.899999976158142,0.530611963064732)
--(axis cs:1,0.525029253160796)
--(axis cs:1,0.506635128716873)
--(axis cs:1,0.506635128716873)
--(axis cs:0.899999976158142,0.505749360065299)
--(axis cs:0.800000011920929,0.507601320505365)
--(axis cs:0.699999988079071,0.509059868166455)
--(axis cs:0.600000023841858,0.503590934798443)
--(axis cs:0.5,0.503382422930687)
--(axis cs:0.400000005960464,0.503536945914927)
--(axis cs:0.300000011920929,0.504408002944745)
--(axis cs:0.200000002980232,0.498958839692707)
--(axis cs:0.100000001490116,0.498241410574627)
--(axis cs:0,0.493694824786895)
--cycle;

\addplot [line width=0.7pt, steelblue76114176]
table {%
0 0.503916189795646
0.100000001490116 0.508825219041835
0.200000002980232 0.508082540171148
0.300000011920929 0.509222623165781
0.400000005960464 0.512036339205987
0.5 0.513916241559677
0.600000023841858 0.51424194053203
0.699999988079071 0.516584609408317
0.800000011920929 0.516377618353513
0.899999976158142 0.516196897547774
1 0.516570308289621
};
\addlegendentry{LGBMRegressor}
\end{axis}

\end{tikzpicture}

%% file: figures/noise_sweep/EAF_Model_LGBMRegressor_LIME_free.tex
\begin{tikzpicture}[
scale=\xaifigscale,
every axis/.style={xtick distance=0.25, legend pos=south west, legend style={
                           font=\Large
                },
            restrict y to domain=0:1
                }
]

\definecolor{darkslategray38}{RGB}{38,38,38}
\definecolor{lavender234234242}{RGB}{234,234,242}
\definecolor{steelblue76114176}{RGB}{76,114,176}

\begin{axis}[
axis background/.style={fill=lavender234234242},
axis line style={white},
legend cell align={left},
legend style={fill opacity=0.8, draw opacity=1, text opacity=1, draw=none, fill=lavender234234242},
tick align=outside,
tick pos=left,
x grid style={white},
xlabel=\textcolor{darkslategray38}{Noise Level},
xmajorgrids,
xmin=-0.05, xmax=1.05,
xtick style={color=darkslategray38},
xtick={-0.2,0,0.2,0.4,0.6,0.8,1,1.2},
xticklabels={
  \(\displaystyle {\ensuremath{-}0.2}\),
  \(\displaystyle {0.0}\),
  \(\displaystyle {0.2}\),
  \(\displaystyle {0.4}\),
  \(\displaystyle {0.6}\),
  \(\displaystyle {0.8}\),
  \(\displaystyle {1.0}\),
  \(\displaystyle {1.2}\)
},
y grid style={white},
ylabel=\textcolor{darkslategray38}{Score \(\displaystyle s\)},
ymajorgrids,
ymin=0, ymax=1,
ytick style={color=darkslategray38},
ytick={0,0.2,0.4,0.6,0.8,1},
yticklabels={
  \(\displaystyle {0.0}\),
  \(\displaystyle {0.2}\),
  \(\displaystyle {0.4}\),
  \(\displaystyle {0.6}\),
  \(\displaystyle {0.8}\),
  \(\displaystyle {1.0}\)
}
]
\path [draw=steelblue76114176, fill=steelblue76114176, opacity=0.3, line width=0.12pt]
(axis cs:0,0.489496529102325)
--(axis cs:0,0.499418199062347)
--(axis cs:0.100000001490116,0.508954519033432)
--(axis cs:0.200000002980232,0.503887689113617)
--(axis cs:0.300000011920929,0.503619647026062)
--(axis cs:0.400000005960464,0.50151926279068)
--(axis cs:0.5,0.504734462499619)
--(axis cs:0.600000023841858,0.500956416130066)
--(axis cs:0.699999988079071,0.504536366462707)
--(axis cs:0.800000011920929,0.506712830066681)
--(axis cs:0.899999976158142,0.506563687324524)
--(axis cs:1,0.507955312728882)
--(axis cs:1,0.489709854125977)
--(axis cs:1,0.489709854125977)
--(axis cs:0.899999976158142,0.48415242433548)
--(axis cs:0.800000011920929,0.489607042074203)
--(axis cs:0.699999988079071,0.493694120645523)
--(axis cs:0.600000023841858,0.490598732233048)
--(axis cs:0.5,0.490648609399796)
--(axis cs:0.400000005960464,0.487610465288162)
--(axis cs:0.300000011920929,0.489197653532028)
--(axis cs:0.200000002980232,0.491742807626724)
--(axis cs:0.100000001490116,0.502375429868698)
--(axis cs:0,0.489496529102325)
--cycle;

\addplot [line width=0.7pt, steelblue76114176]
table {%
0 0.494711066995348
0.100000001490116 0.505221173167229
0.200000002980232 0.49724495112896
0.300000011920929 0.494669198989868
0.400000005960464 0.494454550743103
0.5 0.496352446079254
0.600000023841858 0.496655878424644
0.699999988079071 0.497453796863556
0.800000011920929 0.497699308395386
0.899999976158142 0.497852912545204
1 0.500632303953171
};
\addlegendentry{LGBMRegressor}
\end{axis}

\end{tikzpicture}

%% file: tables/sanity_xai_val.tex
\begin{tabular}{rllllll}
\toprule
Model & $s_{\text{XAI}}$ Grad & $s_{\text{XAI}}$ \gls{sg} & $s_{\text{XAI}}$ \gls{ale}-\gls{knn} & $s_{\text{XAI}}$ \gls{lime} & $s_{\text{XAI}}$ \gls{shap} \\
\midrule
LGBMRegressor & 0.50$\pm$0.02 & 0.52$\pm$0.04 & 0.54$\pm$0.04 & 0.49$\pm$0.02 & 0.50$\pm$0.03 \\
Linear Regression & 0.57$\pm$0.03 & 0.57$\pm$0.03 & 0.57$\pm$0.03 & 0.51$\pm$0.02 & 0.49$\pm$0.03 \\
Neural Network Ensemble & 0.52$\pm$0.03 & 0.53$\pm$0.03 & 0.54$\pm$0.04 & 0.50$\pm$0.02 & 0.50$\pm$0.03 \\
\bottomrule
\end{tabular}

%% file: tables/sanity_xai_train.tex
\begin{tabular}{rllllll}
\toprule
Model & $s_{\text{XAI}}$ Grad & $s_{\text{XAI}}$ \gls{sg} & $s_{\text{XAI}}$ \gls{ale}-\gls{knn} & $s_{\text{XAI}}$ \gls{lime} & $s_{\text{XAI}}$ \gls{shap} \\
\midrule
LGBMRegressor & 0.50$\pm$0.02 & 0.48$\pm$0.07 & 0.48$\pm$0.13 & 0.49$\pm$0.10 & 0.50$\pm$0.04 \\
Linear Regression & 0.51$\pm$0.10 & 0.51$\pm$0.10 & 0.51$\pm$0.10 & 0.47$\pm$0.16 & 0.45$\pm$0.24 \\
Neural Network Ensemble & 0.74$\pm$0.16 & 0.75$\pm$0.17 & 0.75$\pm$0.17 & 0.49$\pm$0.11 & 0.47$\pm$0.11 \\
\bottomrule
\end{tabular}